\PassOptionsToPackage{table}{xcolor}

\documentclass[acmsmall]{acmart}
\makeatletter
\renewcommand\normalsize{%
   \@setfontsize\normalsize{9}{11pt} 
   \abovedisplayskip 6pt plus 2pt minus 2pt
   \belowdisplayskip \abovedisplayskip
   \abovedisplayshortskip 0pt plus 2pt
   \belowdisplayshortskip 3pt plus 2pt minus 2pt
}
\makeatother

\usepackage[table]{xcolor} 
\usepackage{longtable} 
\usepackage{rotating}
\usepackage{float}
\usepackage{xcolor}
\usepackage{tabularx}
\usepackage{makecell}
\usepackage{geometry}
\usepackage{float}
\usepackage{multirow}
\usepackage{graphicx}
\usepackage{subcaption}
\usepackage{lscape}
\usepackage{rotating}
\usepackage{graphicx}
\usepackage{adjustbox}

\begin{document}

\title{Exploring Graph Mamba: A Comprehensive Survey on State-Space Models for Graph Learning}


\author{Safa Ben Atitallah}
\email{satitallah@psu.edu.sa}
\affiliation{%
  \institution{Prince Sultan University}
  \city{Riyadh}
  \country{Saudi Arabia}
}
\affiliation{%
  \institution{University of Manouba}
  \city{Manouba}
  \country{Tunisia}
}

\author{Chaima Ben Rabah}
\affiliation{%
  \institution{Weill Cornell Medicine}
  \city{Doha}
  \country{Qatar}
}
\affiliation{%
  \institution{University of Manouba}
  \city{Manouba}
  \country{Tunisia}
}

\author{Maha Driss}
 \affiliation{%
  \institution{Prince Sultan University}
  \city{Riyadh}
  \country{Saudi Arabia}
}
\affiliation{%
  \institution{University of Manouba}
  \city{Manouba}
  \country{Tunisia}
}

\author{Wadii Boulila}
 \affiliation{%
  \institution{Prince Sultan University}
  \city{Riyadh}
  \country{Saudi Arabia}
}
\affiliation{%
  \institution{University of Manouba}
  \city{Manouba}
  \country{Tunisia}
}

\author{Anis Koubaa}
 \affiliation{%
  \institution{Prince Sultan University}
  \city{Riyadh}
  \country{Saudi Arabia}
}

\renewcommand{\shortauthors}{Ben Atitallah et al.}

\begin{abstract}
  Graph Mamba, a powerful graph embedding technique, has emerged as a cornerstone in various domains, including bioinformatics, social networks, and recommendation systems. This survey represents the first comprehensive study devoted to Graph Mamba, to address the critical gaps in understanding its applications, challenges, and future potential. We start by offering a detailed explanation of the original Graph Mamba architecture, highlighting its key components and underlying mechanisms. Subsequently, we explore the most recent modifications and enhancements proposed to improve its performance and applicability. To demonstrate the versatility of Graph Mamba, we examine its applications across diverse domains. A comparative analysis of Graph Mamba and its variants is conducted to shed light on their unique characteristics and potential use cases. Furthermore, we identify potential areas where Graph Mamba can be applied in the future, highlighting its potential to revolutionize data analysis in these fields. Finally, we address the current limitations and open research questions associated with Graph Mamba. By acknowledging these challenges, we aim to stimulate further research and development in this promising area. This survey serves as a valuable resource for both newcomers and experienced researchers seeking to understand and leverage the power of Graph Mamba.
\end{abstract}

\keywords{State Space Models,
 Mamba Block,
Graph Mamba,
Graph Learning,
Graph Convolutional Network,
Applications}


\maketitle

\section{Introduction}

Graph-based learning models, particularly Graph Neural Networks (GNNs), have gained significant traction in recent years due to their ability to effectively capture and process complex relational data. These models have proven advantageous in many different fields where graphs are the typical way to represent data \cite{xia2021graph}.
The increasing significance of GNNs can be attributed to various factors. Graph-structured data has been raised in many real-world systems, such as social networks, molecular structures, and citation networks \cite{jiang2022graph,wang2024graph}. GNNs have a solid ability to leverage relational information and the connections between entities. In addition, different advanced GNN architectures have been proposed with high scalability to handle large-scale graphs, making them suitable for big data applications. This type of learning can be applied to various tasks, including node classification, link prediction, and graph classification.

However, they face several significant challenges that limit their effectiveness in specific scenarios. Most GNNs are restricted in their ability to effectively capture long-range dependencies. They typically rely on message passing between neighboring nodes, which can lead to information dilution over multiple hops. This constraint is particularly problematic in graphs with complex hierarchical structures. 
In addition, many GNN architectures require multiple rounds of neighborhood aggregation, which is computationally expensive, especially for large-scale graphs. The computational cost grows significantly as the number of layers increases to capture more complex patterns.
Furthermore, GNNs usually face memory constraints and increased training time when applied to large graphs \cite{liu2024federated}. The issue is heightened for dynamic graphs, where the structure changes over time and requires frequent updates to node representations. Sampling techniques have been proposed to address this but can lead to information loss. GNN variants have quadratic complexity regarding the number of nodes or tokens. Similar issues arise in GNNs when computing full graph attention or when dealing with dense graphs. This quadratic scaling significantly impacts performance and limits the application of these models to huge graphs or long sequences. 

Indeed, addressing the limitations of current graph-based learning models is crucial for their broader applicability. One promising direction in this effort is the adaptation of State-Space Models (SSMs) to graph learning, which has led to the development of Graph Mamba. SSMs are mathematical models initially designed for sequence modeling in control theory and signal processing. They represent a system's behavior using a set of input, output, and state variables related by first-order differential equations. In the context of ML, SSMs can efficiently model long-range dependencies in sequential data. They offer a continuous-time perspective on sequence modeling, which can benefit specific data types.

Recently, Mamba has emerged as a groundbreaking approach in Artificial Intelligence (AI), specifically designed as a specialized form of the SSM to address the computational limitations of traditional Deep Learning (DL) models. Standard models, such as Convolutional Neural Networks (CNNs) and Transformers, face a significant challenge related to computational inefficiency, particularly in tasks involving long-sequence modeling. Mamba's primary goal is to enhance computational efficiency by reducing time complexity from quadratic, as seen in transformers, to linear. Inspired by advancements in structured SSMs, Mamba is presented to boost performance in areas requiring long-range dependency modeling and large-scale data processing.

Graph Mamba emerges as a specialized variant of SSMs designed specifically for graph learning. Its primary goal is to address the limitations of traditional GNNs by leveraging the unique strengths of state-space models. The core concept of Graph Mamba is its state-space modeling approach, which employs selective scanning, a powerful mechanism for efficiently processing graph information by dynamically focusing on the most relevant parts of the graph structure. This allows Graph Mamba to manage large-scale and complex graphs with superior computational performance.
 
Recently, there has been an increasing interest in Graph Mamba, as shown by the growing number of articles. 
This survey aims to investigate the potential of integrating graph structures with Mamba frameworks to enhance representation learning and scalability.  Through a comparative analysis of existing literature and empirical studies, this survey evaluates the performance of Graph Mamba against traditional Machine Learning (ML) methods.


\subsection{Related Surveys}
This section provides a thorough summary of essential survey studies from two research fields; GNN architectures and Mamba framework. 
\subsubsection{Surveys on Graph Neural Networks: Advancements and Applications}

GNNs have found applications in a variety of domains, including computer vision, recommendation systems, fraud detection, and healthcare. Several comprehensive surveys have been elaborated on GNNs.
In \cite{khemani2024review}, the authors presented a comprehensive review of GNNs, emphasizing their evolution, essential concepts, and numerous potential applications of this cutting-edge technology. GNNs transformed ML by effectively modeling relationships in graph-structured data, overcoming the constraints of conventional neural networks. The study described major GNN architectures such as Graph Convolutional Networks (GCNs), Graph Attention Networks (GATs), and Graph Sample and Aggregate (GraphSAGE), as well as their message-passing mechanisms for repeatedly aggregating information from neighboring nodes. Among the applications investigated were node classification, link prediction, and graph classification across social networks, biology, and recommendation systems. In addition, the paper examined commonly used datasets and Python libraries, explored scalability and interpretability issues, and recommended future research areas to improve GNN performance and expand its applicability to dynamic and heterogeneous graphs.
The authors in \cite{wu2020comprehensive} provided a comprehensive review of GNNs and their applications in data mining and ML fields. It discussed the issues posed by graph-structured data in non-Euclidean domains and how DL methods had been modified to accommodate such data. The authors in \cite{wu2020comprehensive} presented a new taxonomy that divided GNNs into four types: recurrent GNNs, convolutional GNNs, graph autoencoders, and spatial-temporal GNNs, each of which was customized to a specific graph-based task. The survey also examined the practical uses of GNNs in social networks, recommendation systems, and biological modeling. Furthermore, it reviewed open-source implementations, benchmark datasets, and evaluation criteria utilized in GNN research. It concluded by listing unresolved challenges and proposing future research topics, highlighting the potential for advanced GNN methodologies and applications.

\subsubsection{Surveys on Mamba: Trends, Techniques, and Applications}

Since its introduction in late 2023, Mamba has received a lot of attention in the DL community because it offers compelling benefits that encourage adoption and exploration across multiple domains. Nnumerous surveys have been elaborated to investigate 
MAmba potential and its applications.
For example, The Patro \textit{et al.} in \cite{patro2024mamba} investigated the use of SSMs as efficient alternatives to transformers for sequence modeling applications. It classified SSMs into three paradigms: gating, structural, and recurrent, and discussed key models like S4, HiPPO, and Mamba. This survey emphasized the use of SSMs in a variety of domains, including natural language processing, vision, audio, and medical diagnostics. It compared SSMs and transformers based on computational efficiency and benchmark performance. 
The paper emphasized the need for additional research to improve SSMs' ability to handle extended sequences while maintaining high performance across multiple applications.

Qu \textit{et al.} \cite{qu2024survey} gave a thorough explanation of Mamba. They positioned Mamba as a viable alternative to transformer topologies, particularly for tasks involving extended sequences. The survey presents the fundamentals of Mamba, highlighting its incorporation of features from RNNs, Transformers, and SSMs. It examined improvements in Mamba design, including the creation of Mamba-1 and Mamba-2, which featured breakthroughs such as selective state space modeling, HiPPO-based memory initialization, and hardware-aware computation optimization methods. The authors also looked into Mamba's applications in a variety of domains, including natural language processing, computer vision, time-series analysis, and speech processing, demonstrating its versatility in tasks such as large language modeling, video analysis, and medical imaging. The study identified many problems related to Mamba use, including limitations in context-aware modeling and trade-offs between efficiency and generalization. They also suggested improvements for Mamba's generalization capabilities, computational efficiency, and discussed its applicability in new research areas in the future.

In their recent study, Wang \textit{et al.} in \cite{wang2024state}  conducted a comprehensive survey that emphasized the changing landscape of DL technologies. This survey focused primarily on the theoretical foundations and applications of SSMs in fields such as natural language processing, computer vision, and multi-modal learning, with the goal of addressing the computational inefficiencies of conventional models. Experimental comparisons revealed that, while SSMs showed promise in terms of efficiency, they frequently fell short of the performance of cutting-edge transformer models. Despite this, the findings in this study revealed that SSMs could reduce memory usage and provide insights into future research to improve their performance. This study provided valuable insights into DL architectures, showing that SSMs could play a crucial role in their development. 

On the other hand, recent studies have explored Mamba Vision techniques, emphasizing its rapid growth and rising importance in computer vision. They highlight Mamba's ability to address the limitations of CNNs and Vision Transformers, particularly in capturing long-range dependencies with linear computational complexity. Rahman \textit{et al.} \cite{rahman2024mamba} investigated the Mamba model, this revolutionary computer vision approach that addressed the constraints of CNNs and Vision Transformers (ViTs). With CNNs, local feature extraction is more efficient, but with ViTs, long-range dependencies are more difficult due to their quadratic self-attention mechanism. Mamba used Selective Structured State Space Models (S4) to handle long-range dependencies with linear computational cost-efficiently. The survey classified Mamba models into four application categories: video processing, medical imaging, remote sensing, and 3D point cloud analysis. A variety of scanning approaches were also examined, including zigzag, spiral, and omnidirectional methods. The paper emphasized Mamba's computational efficiency and scalability, which make it suitable for high-resolution and real-time operations. The authors also conducted a comparison investigation of Mamba against CNNs and ViTs, proving its advantages in a variety of benchmarks. They also discussed potential future research directions, such as increasing dynamic state representations and model interpretability. Overall, the article positioned Mamba as a paradigm for balancing performance and computation efficiency in computer vision.

The study presented in \cite{liu2024vision} provided a comprehensive survey and taxonomy of SSMs in vision-oriented approaches, with a focus on Mamba. A comparison was made between Mamba, CNNs, and Transformers. Due to its ability to handle irregular and sparse data, Mamba has been used for a variety of vision applications, including medical image analysis, remote sensing, and 3D visual identification. This survey classified Mamba models by application areas, such as general vision, multi-modal tasks, and vertical-domain tasks, and presented a comprehensive taxonomy of Mamba variants, as well as detailed descriptions of their principles and applications. The main objective of this survey was to help academics comprehend Mamba's development and potential to improve computer vision, particularly in applications that require computing efficiency and long-range dependency modeling.

\subsubsection{Discussion}

While the surveys discussed above provide essential insights into a variety of cutting-edge fields, they do have significant limitations. Many surveys on GNNs concentrate on the theoretical foundations and architecture of these networks, paying little attention to practical problems and model scalability in dynamic scenarios. In addition, while these surveys highlight GNN's relevance in research fields like healthcare and recommendation systems, they often ignore practical challenges such as computational complexity, scalability in large networks, and limited generalization across heterogeneous datasets. Besides, while many surveys discuss Mamba frameworks' potential to overcome transformer limitations, they tend to focus on theoretical advancements and model efficiency rather than providing an in-depth analysis of real-world limitations, such as trade-offs between computational efficiency and performance across various domains. 
The available studies on GNNs and Mamba models highlight their distinct improvements but remain limited in scope. GNN surveys investigate graph-based learning but do not explore how graph structures may be incorporated into Mamba frameworks. Mamba-related surveys, on the other hand, concentrate on sequential modeling and computing efficiency without investigating the possibility of combining graph-based methods. This discrepancy creates a huge research gap. Integrating graph structures into Mamba presents transformative capabilities that need a comprehensive review. 

\subsection{Contributions of the Proposed Survey}

There has been a rapid surge in research exploring Graph Mamba's architecture, improvements, and applications across various domains. However, the insights remain distributed across various studies, and there is currently no thorough review that brings these findings together. As the field advances rapidly, a well-structured overview of the latest developments is increasingly valuable. The main contributions of this survey paper are illustrated in the following points:

\begin{itemize}
    
    \item This survey offers a comprehensive explanation of the fundamental principles of Graph Mamba and offers a strong theoretical foundation for both researchers and practitioners.

    \item It examines the most recent enhancements to the original Graph Mamba architecture and evaluates the performance implications of various proposed modifications.
    
    \item A comparison of various Graph Mamba variants is presented to emphasize their unique characteristics.

    \item The survey examines a variety of disciplines in which Graph Mamba has been implemented, such as computer vision, healthcare, and biosignals.

    \item Additionally, it identifies potential fields for future implementations of Graph Mamba and addresses the current limitations and open research questions in this context.

\end{itemize}

\subsection{Paper Organization}
This survey provides a comprehensive overview of Graph Mamba state space models, including their architectures, applications, challenges, and potential future directions. We explore the advantages and disadvantages of existing Graph Mamba models and discuss their prospects for future development. The paper is organized as follows: Section~\ref{sec2} discusses the preliminaries and key terms related to Graph Neural Networks, State Space Models, and Mamba. In Section~\ref{sec3}, we delve into various Graph Mamba architectures. Section~\ref{sec4} highlights recent applications of Graph Mamba. Sections~\ref{sec5} and \ref{Section6new}  present benchmarks and a comparative analysis of results demonstrating Graph Mamba's performance across different tasks. 
Section~\ref{sec6} outlines the limitations of applying Graph Mamba. Section~\ref{sec7} explores emerging areas and future research directions. Finally, we conclude the work in Section~\ref{sec8}.


\section{Preliminaries}
\label{sec2}
This section reviews the foundation of GNNs and SSMs and how they are integrated in the Graph Mamba framework.

\subsection{Graph Neural Networks (GNNs)}

GNNs have developed as a strong class of DL models built for graph-structured data. Unlike standard ML models, which often operate on fixed-sized inputs such as pictures or sequences, GNNs are specially designed to handle non-Euclidean data, represented as nodes and edges \cite{xia2021graph}. This makes GNNs ideal for tasks that need complicated relational data, such as social networks, knowledge graphs, chemical structures, and recommendation systems. Graphs are inherently adaptable and can represent a broad range of data formats. Standard DL models, such as CNNs, perform well with structured data like grids or sequences but fail to generalize to graph data. GNNs address this drawback by learning representations of nodes, edges, and graphs in a way that captures both the local neighborhood information and the global structure of the graph. Indeed, GNNs are based on the idea of message forwarding, in which each node in the network gathers information from its neighbors to update its representation. This method enables GNNs to effectively capture both local patterns and long-range relationships throughout the graph by propagating information through a set of layers. In the following subsections, we present an overview about some popular GNN architectures proposed in the literature.

\subsubsection{Graph Convolutional Networks (GCNs)}

GCNs, introduced by Kipf \textit{et al.} \cite{kipf2016semi}, are a specialized type of GNN created to work with graph-based data. The core idea is to take the concept of convolution, which is so effective in image processing with grids of pixels, and adapt it to the irregular structure of graphs. In contrast to conventional CNNs that depend on static grids, GCNs execute localized convolutions at each node, aggregating information from adjacent nodes. This enables GCNs to understand the links and patterns inside the graph structure in a manner that conventional CNNs can't. The propagation rule for a GCN layer is represented as:

\begin{equation}
    h_i^{(l+1)} = \sigma\left(\sum_{j\in \mathcal{N}(i)} \frac{1}{c_{ij}} W^{(l)}h_j^{(l)}\right)
\end{equation}

where $i$ is the node being processed, $N_{i}$ is the set of nodes that are neighbors of $i$, $h_{i}^{l}$ is the mathematical representation of $i$ at layer $l$, $W^{l}$ is the layer's weight matrix, and $c_{ij}$ serves as a normalization factor to account for differences in the number of neighbors.

The graph convolution process is carried out repeatedly on many levels, which helps the model understand more complicated connections and higher-level links in the graph structure.


\subsubsection{Graph Attention Networks(GATs)}

In \cite{hamilton2017inductive}, GATs have been proposed by Velikovi \textit{et al.}, which are a type of GNN designed to address limitations in traditional GNNs. They are specially designed for complex connections and irregular graph structures. Their key innovation is an attention mechanism that selectively aggregates information from neighboring nodes, allowing them to focus on the most relevant inputs. This method assigns different weights to each neighbor, emphasizing the importance of specific nodes during aggregation and improving the model's ability to capture meaningful relationships. The computations made in the GAT layer are presented in the following Equation \ref{eq:gat}:
\begin{equation}
    h_i^{(l+1)} = \sigma\left(\sum_{j\in \mathcal{N}(i)} \alpha_{ij} W^{(l)}h_j^{(l)}\right)
\label{eq:gat}
\end{equation}

where, \( i \) denotes the target node, \( N_{(i)} \) represents the set of \( i \)’s neighbors, and \( h_i^{(l)} \) is the representation of node \( i \) at layer \( l \). \( W^{(l)} \) is the weight matrix shared across layer \( l \), and \( \alpha_{ij} \) is the attention weight for the edge between nodes \( i \) and \( j \), determined by a learnable attention mechanism.

\subsubsection{Graph Sample and Aggregation (GraphSAGE)}

GraphSAGE, introduced by Hamilton \textit{et al.} in \cite{hamilton2017inductive}, is a scalable GNN architecture for large graphs. It learns node embeddings by sampling and aggregating information from local neighbors, allowing inductive learning to generalize to unseen nodes. GraphSAGE consists of two main parts: embedding generation (forward propagation) and parameter learning. The model iteratively traverses neighborhood layers and enables nodes to gather information from their surroundings. The representation for a node $v$ at depth $k$ is updated as follows:

\begin{equation}
h_v^{(k)} = \sigma\left(W_k \cdot {CONCAT}\left(h_v^{(k-1)}, AGGREGATE_k\left(\{h_u^{(k-1)}, \forall u \in N(v)\}\right)\right)\right)
\end{equation}

where, \( \sigma \) is the non-linear activation function, and \( W_k \) is the learnable weight matrix for depth \( k \). The \text{CONCAT} operation combines \( h_v^{(k-1)} \) with the aggregated data from \( v \)'s neighbors, denoted \( N(v) \), using \( \text{AGGREGATE}_k \), which can be a mean, LSTM, or pooling function. This iterative process enables GraphSAGE to capture complex node relationships in an inductive and scalable way.

\subsection{State Space Models (SSMs)}

DL has seen a notable transformation with the emergence of Transformer models, which have attained dominance in both computer vision and natural language processing. Their success is attributed to the self-attention mechanism, an effective strategy that enhances model understanding by producing an attention matrix based on query, key, and value vectors \cite{boulila2024transformer}. This methodology has transformed how models analyze and comprehend data. However, the Transformer architecture faces a notable challenge. Its self-attention mechanism operates with quadratic time complexity. As the input sequence length grows, the computational requirements increase exponentially and create a significant bottleneck, especially when dealing with very long sequences or large datasets. This limitation has pushed research to develop more efficient architectures that can maintain the benefits of self-attention while scaling more effectively to more significant inputs.

In this context, Mamba was proposed by Gu \textit{et al.} \cite{gu2023mamba} based on SSMs. It has gained much interest in recent years due to its effectiveness in providing good performance as transformers while reducing the overall complexity. SSMs are widely used to represent dynamic systems \cite{hamilton1994state}. They convert a one-dimensional input sequence $(u(t))$ into an N-dimensional continuous latent state $(x(t))$ and project it into a one-dimensional output signal $(y(t))$. Equations \ref{eq:1} and \ref{eq:2} describe this transformation process:

\begin{equation}
    x'(t) = A x(t) + B u(t)
    \label{eq:1}
\end{equation}

\begin{equation}
    y(t) = C x(t) + D u(t)
    \label{eq:2}
\end{equation}

where \( A \) represents the system's dynamics matrix, \( B \) refers to the input transformation, \( C \) projects the latent state \( x(t) \) into the output space, and \( D \) provides a direct input-to-output mapping. These parameters are initialized differently depending on specific SSM variants, such as S4, S5, and S6. These SSM variants are illustrated in the following subsections. 



\subsubsection{Structured State Space Sequence Models (S4)}
S4 was introduced to address the inefficiencies of traditional transformers in sequence modeling \cite{gu2021efficiently}. It leverages structured matrices that allow for the fast and efficient modeling of long sequences. S4 transforms the input sequences into latent states that evolve over time using a continuous-state model. This method is particularly suited for handling very long-range dependencies, as it avoids the quadratic complexity of self-attention mechanisms.

\subsubsection{Simplified State Space Layers for Sequence Modeling (S5)}
S5 builds on the foundation of S4 but simplifies the model architecture for more efficient sequence modeling \cite{smith2022simplified}. It reduces the complexity of the latent state transformation process, making it faster and more scalable while retaining the ability to model long-range dependencies.

\subsubsection{Selective State Space Models (S6)}

S6 is the most recent development in SSMs and has been widely adopted, particularly in graph-based learning tasks like Graph Mamba \cite{gu2021efficiently}. The introduction of a selective scanning mechanism improves its capacity to efficiently manage both complex graph structures and long sequences. 
In this survey, we provide a comprehensive overview of S6, detailing its architecture and the contributions of the selective scanning mechanism.

The selective scanning mechanism in S6 is designed to overcome the limitations faced by standard sequence models. In contrast to conventional models that treat all segments of the input sequence equally, S6 selectively identifies and emphasizes the most relevant portions of the input. This allows the model to concentrate on relevant information and minimize the cost of computing linked to processing irrelevant data points. The selective scanning approach significantly decreases temporal complexity, making S6 scalable for practical applications. The equations below present the foundation of the S6 model.

\begin{itemize}
    \item \textbf{State Transition with Selection:}
    \begin{equation}
        B(L, N) \leftarrow s_B(x)
    \end{equation}

    The function \( s_B(x) \) introduces selectivity in the state transition process by dynamically adapting the state matrix \( B \) based on input \( x \). This selective focus enables the model to emphasize certain state components that are contextually important, hence enhancing the importance of state transitions within the sequence.
    
    \item \textbf{Output Mapping with Selection:}
    
    \begin{equation}      
        C(L, N) \leftarrow s_C(x)
    \end{equation}
    
    Where \( s_C(x) \) acts as a filtering mechanism in the output mapping process, directing attention to particular components of \( x \) that contribute most meaningfully to the final output. By selectively mapping relevant features, the model reduces noise and enhances the clarity of information passed to subsequent layers. 

    \item \textbf{Parameter Adjustment with Selection:}
    \begin{equation}
        \Delta (L, D) \leftarrow \tau(\text{Parameter} + s_\Delta(x))
    \end{equation}
    
    In this step, \( s_\Delta(x) \) dynamically adjusts the parameters within \(\Delta\) in response to variations in the input $x$. The function $\tau$ then consolidates these adjustments. This adaptability enhances the model’s capability to generalize across diverse inputs 

    \item \textbf{Discrete State Evolution:}
    \begin{equation}
        \bar{A}, \bar{B} \leftarrow \text{discretize}(\Delta, A, B)
    \end{equation}
    
    The discretization technique converts continuous model dynamics into a discrete representation. This makes them easier to process linearly and saves a lot of computing power. By updating the matrices \( \Delta \), \( A \), and \( B \), this step enables the model to preserve essential temporal relationships within the data while optimizing for discrete operations.

    \item \textbf{Final Output:}
    \begin{equation}
        y \leftarrow \text{SSM}(\bar{A}, \bar{B}, C)(x)
    \end{equation}

    The final output \( y \) is generated by the SSM, which uses the selectively updated matrices \( \bar{A} \), \( \bar{B} \), and \( C \). This process takes the selectively processed input data and turns it into an improved output that captures the structure and relationships of the input data properly. 
    
\end{itemize}

In the context of graph-based learning, S6 has proven particularly useful. Graph data often includes complex relationships between nodes and edges, and processing every node in a large graph can become computationally prohibitive. S6's selective scanning mechanism allows the model to concentrate on the most critical substructures of the graph. This is vital for applications that require long-range dependencies or deep graph traversal. 





\subsection{Transition from Graph Learning and SSMs to Graph Mamba}

While standard SSMs were built to handle sequence data, many real-world tasks need graph-structured data, in which nodes represent entities and edges describe relationships. Unlike sequences, graphs are non-Euclidean, which means they lack a natural order. This raises the challenge of the complex modeling of long-range connections and multi-hop interactions between nodes. To expand the capabilities of SSMs to graphs, Graph Mamba was designed as a specialized variation of SSMs for graph learning. This design uses state-space modeling approaches to address the specific problems of graph-based tasks. By leveraging the linear complexity of SSMs, Graph Mamba provides a more efficient alternative to GNNs. The main elements of this transition include:

\begin{itemize}
    \item \textbf{Linear Complexity:} \\ One of the standout features of Graph Mamba is its inherited linear time complexity from SSMs, making it significantly more efficient compared to standard GNNs. Graph Mamba can analyze long-range relationships via linear scaling, which improves speed on big networks.

    \item \textbf{Unified Spatial and Temporal Processing:} \\ Graph Mamba is designed to handle both spatial (node and edge relationships) and temporal (changes over time) information in a unified framework. This makes it especially useful for dynamic graph workloads like growing social networks or temporal knowledge graphs.

    \item \textbf{Selective Scanning Mechanisms:}\\ Graph Mamba, like SSMs, uses selective scanning techniques to effectively process graph information and use the structure's most important components. This enables it to capture key information while minimizing unnecessary computations.
\end{itemize}
 
\section{Graph Mamba }
\label{sec3}
In this section, we provide an overview of Graph Mamba architecture. In addition, we discuss the graph structures used, the selective scanning methods designed, and the different training strategies followed for Graph Mamba.

\subsection{Graph Mamba Architecture}

\begin{figure}[h!]
    \centering
    \includegraphics[width=0.9\linewidth]{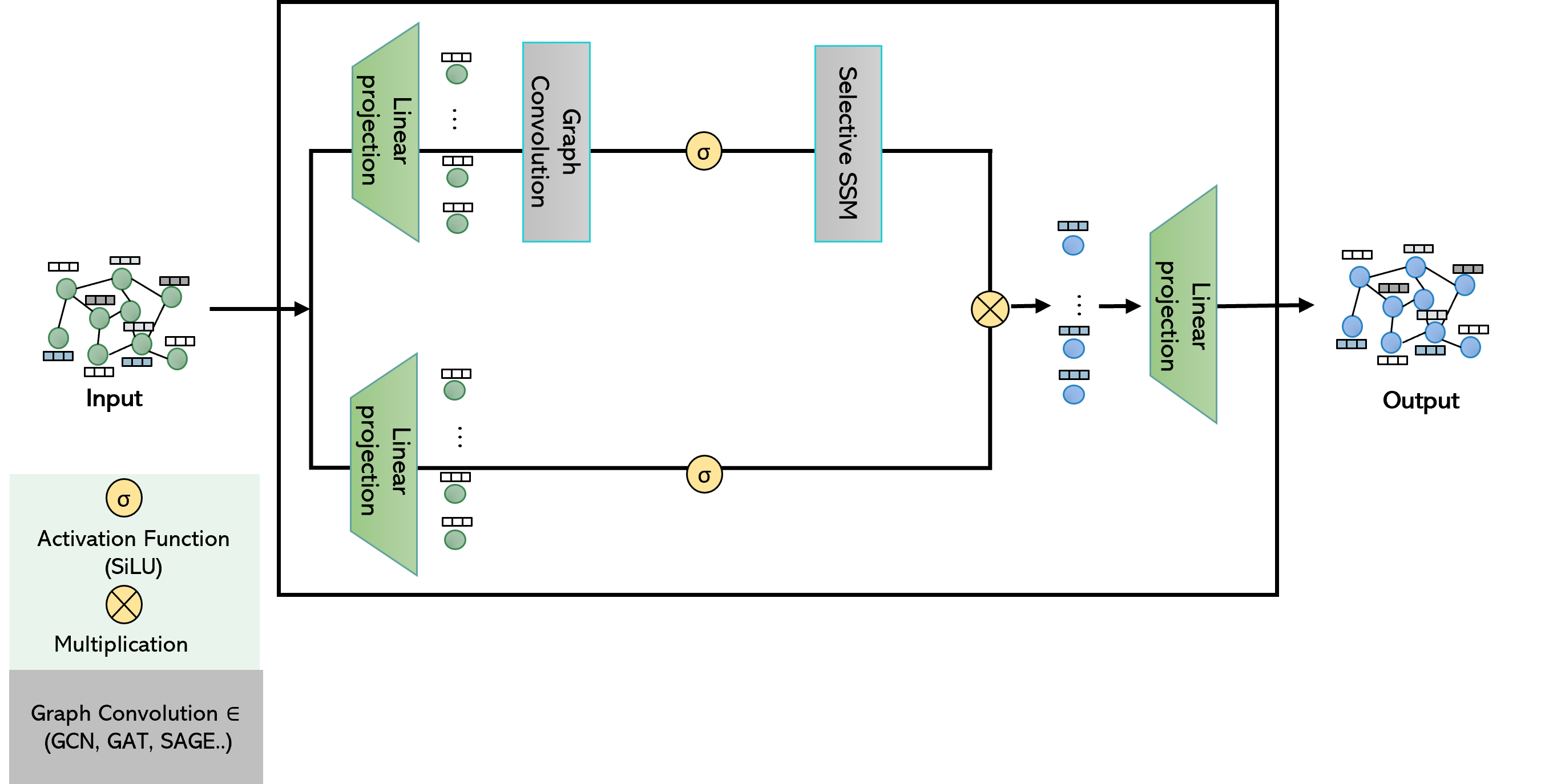}
    \caption{Graph Mamba Architecture: including sequence transformations (Conv and SSM blocks), linear projections, and non-linearity through activation and multiplication operations.}
    \label{fig:Gmamba}
\end{figure}

Graph Mamba is a subset of SSMs created particularly for graph-based learning tasks. Its architecture, as depicted in Figure \ref{fig:Gmamba}, is designed to solve the constraints of classic GNNs, such as computational inefficiency and the difficulty in representing long-range relationships in big networks. 
Graph Mamba's architecture is based on the capacity to represent graphs with linear time complexity, which is more efficient than the quadratic complexity described in many typical GNNs. This is achieved by using SSM concepts that include state updates and latent representations. Graph Mamba recurrently treats data and enables the capture of long-term interactions between nodes in the network without relying on expensive attention techniques like Transformers.

The basic architecture of Graph Mamba consists of three main blocks:

\begin{enumerate}
    \item \textbf{State-Space-Based Message Passing}: Instead of standard message-passing as in GNNs, Graph Mamba uses state-space modeling approaches to transfer information among nodes effectively.
    \item \textbf{Selective Scanning}: Graph Mamba employs a selective mechanism to focus on relevant nodes and edges within the graph. This technique makes it highly efficient when dealing with large graphs where complete message passing is too expensive.
    \item \textbf{Spatial-Temporal Integration}: The model can also handle temporal components of dynamic graphs, making it ideal for tasks requiring both node connections and changes over time.
\end{enumerate}

\subsection{Graph Structure in Mamba Models}

\begin{figure}[h!]
    \centering
    \includegraphics[width=0.7\linewidth]{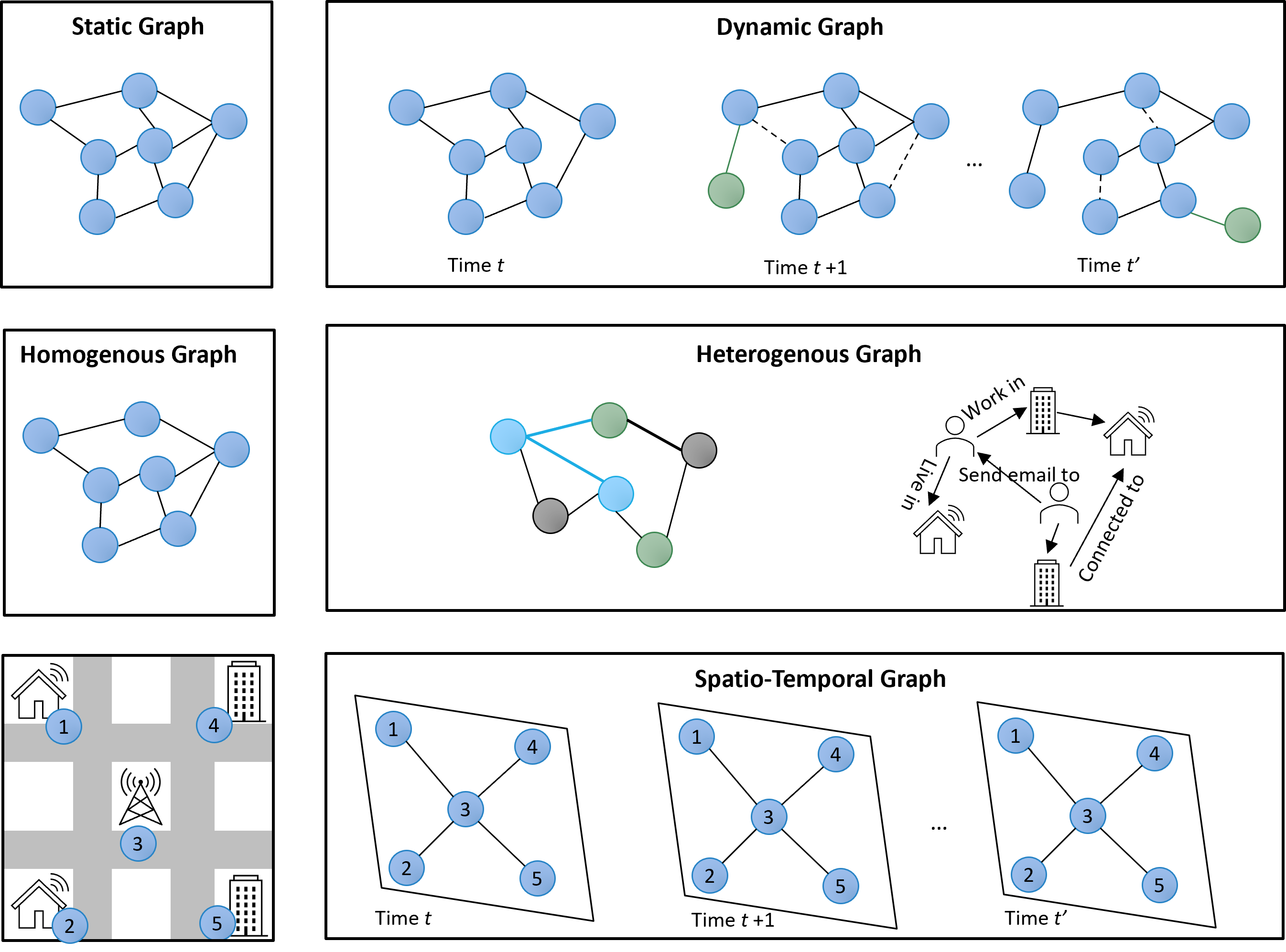}
    \caption{Illustration of different graph types analyzed with the Graph Mamba Framework.}
    \label{fig:G_types}
\end{figure}

The underlying graph structure in the Graph Mamba framework is crucial in shaping the model's data processing and learning potential. Different graph structures capture distinct relationships and dynamics. To maximize its adaptability and effectiveness across diverse applications, we classify these graph structures into three main categories based on the nature of the data they manage; dynamic graph, heterogeneous graph, and spatio-temporal graph. This classification, as presented in Figure \ref{fig:G_types}, allows Graph Mamba to efficiently harness the unique strengths of each graph type. Each category is tailored to specific use cases, allowing the model to address real-world challenges in domains like social networks, biological systems, and temporal-spatial analyses.

\subsubsection{Dynamic Graphs}

Dynamic graphs enhance the Graph Mamba framework's functionality by enabling it to model and learn from developing data structures in which both node attributes and edge connections fluctuate over time. These graphs play an important role in capturing temporal dependencies and adjusting to real-world contexts, where interactions between entities are dynamic and time-sensitive \cite{liang2024survey}.

For instance, STG-Mamba \cite{li2024stg} integrates spatial and temporal relationships to model changing node attributes and structures, excelling in applications like traffic networks where geographical and temporal dynamics are critical. Similarly, continuous-time dynamic graph approaches \cite{li2024dyg, ding2024dygmamba} adapt to scenarios like social networks, where relationships are transient, enabling real-time analysis of evolving interactions. These dynamic graph methodologies demonstrate how evolving graph structures enable deeper insights and robust decision-making in complex and time-sensitive systems.

\subsubsection{Heterogeneous Graph}

Heterogeneous graphs, as used in the Graph Mamba framework, incorporate multiple types of nodes and edges that capture various relationships between entities \cite{bing2023heterogeneous}. Unlike dynamic graphs, which focus on temporal changes, heterogeneous graphs highlight the complexity of node and edge heterogeneity. This graph type allows Graph Mamba to model intricate interactions between different types of entities effectively. In \cite{pan2024hetegraph}, Pan \textit{et al.} introduced the HeteGraph-Mamba, a novel approach to learning with heterogeneous graphs. They use a selective SSM that integrates with the graph learning process. The SSSM improves the handling of heterogeneous data by selectively focusing on important features within the graph's structure. By leveraging this heterogeneous graph structure, Graph Mamba excels in modeling complicated real-world systems. The ability to learn from multiple entities and relationships makes it ideal for applications that need understanding multi-faceted interactions, such as biological system pattern detection.

\subsubsection{Spatio-Temporal Graphs}

Spatio-temporal graphs represent a sophisticated category of dynamic graphs that develop over time and simultaneously enclose spatial relationships among nodes. The graphs illustrate that both nodes and edges, along with their attributes, are affected by spatial proximity and temporal variations \cite{li2024stg}. For example, SpoT-Mamba \cite{choi2024spot} further advances the graph construction by capturing long-range spatial and temporal dependencies using selective state-space modeling.  This enables the model to concurrently learn from the spatial structure and the temporal dynamics of the graph. These graphs are particularly useful for tasks that require understanding both spatial and temporal dependencies, such as traffic flow prediction, social network analysis, and weather forecasting.

\subsection{Selective Scanning Mechanisms for Graph Mamba}

In Graph Mamba, selective scanning is a technique designed to handle graph-structured data in a parallel and recurrent manner. This technique preserves node and edge states that has information from previous stages of the graph traversal. This selective scanning extends the concept of SSMs for graph domains. The main components of this method include:

\begin{itemize}
    \item \textbf{Node and Edge State Update:} The selective scan mechanism changes the state of each node or edge in the graph according to the current input (node/edge characteristics) and the preceding state (data from adjacent nodes/edges). The graph topology determines the flow of information between nodes during this update.
    
    \item \textbf{Selective Propagation:} The selective scan mechanism determines which components of the node/edge states to transmit across the graph and which components to forget or reduce in significance. This selectivity guarantees that only the most useful data from interconnected nodes and edges is retained, while less significant elements are excluded. This is essential for capturing long-range relationships in complex graph architectures.
    
    \item \textbf{Output Generation:} This mechanism produces output for each node and edge after the processing of the chosen and modified states, which are used for further tasks such as node classification, link prediction, or graph classification. 
    
\end{itemize}

Graph Mamba models use various scanning mechanisms to rapidly explore and analyze graph structures, especially in dynamic and spatiotemporal contexts. These techniques help to optimize node and edge selection and traversal, improve computing performance, and ensure accurate graph dependency learning. The following are the primary scanning methods used in Graph Mamba models:
\begin{enumerate}
    \item \textbf{Graph Selective Scan:} \\
    This method selectively incorporates dynamic graph information to optimize which node or edge states are updated and transmitted throughout the graph. It prioritizes the most critical elements of the graph to provide efficient state updates while minimizing computational overhead \cite{han2024innovative} \cite{mehrabian2024mamba}.
    It is used mainly in spatiotemporal models like STG-Mamba \cite{li2024stg}. It helps adapt to changes over time, making it effective for dynamic graphs such as those in transportation, social networks, or communication systems.

    \item \textbf{Temporal Dependency Scan:}\\
    This method can properly capture changes that occur over time by analyzing graph data while considering long-range temporal relationships. It integrates temporal scanning blocks that model time-evolving dependencies in the graph. This approach is widely used in spatiotemporal graph models, such as transportation systems and health data \cite{choi2024spot}.  
     
    \item \textbf{Directed Scan:}\\
    Based on the graph's intrinsic structure, directed scanning processes the graph data in a specific direction. \cite{zhang2024gm}. This method optimizes traversal in graphs where directionality is critical, such as in citation networks or task flow models \cite{zhou2024efficient}.

    \item \textbf{Bi-directional Scanning:} \\
    In this method, nodes are processed both forward and backwards, so graph elements are scanned in two directions instead of just one. The data is processed more thoroughly, and more relationships are extracted, which can be missed in a purely unidirectional scan. Bidirectional scanning is particularly advantageous in tasks like spatial sequence learning, where both past and future states provide crucial context \cite{dong2024hamba}.

    \item \textbf{Recurrent Scan:} \\  
    During a recurrent scan, the model saves context in hidden states and updates output by merging them with the current input. This scanning approach integrates information from earlier phases to improve the learning of sequential or graph data \cite{xu2024identifying,wang2024graph}. Recurrent scans improve computing efficiency by reparameterizing, offering an alternative to costly attention techniques \cite{behrouz2024graph,behrouz2024brain}.
  
    \item  \textbf{Parallel Scanning:}\\
     Parallel scanning improves processing performance by enabling many methods to execute simultaneously instead of sequentially. In the context of Graph Mamba models, this refers to the processing of several portions of the graph in parallel, which speeds up calculation without negatively impacting accuracy. This approach is particularly valuable for large-scale graphs when performance is critical for dealing with massive volumes of data in real-time \cite{yang2024graphmamba} \cite{li2024dyg}.

\end{enumerate}


These scanning methods maximize the exploration and processing of graph structures using Graph Mamba models, especially in dynamic and spatiotemporal settings. Each method contributes to an improved node and edge selection, traversal efficiency, and accurate learning of graph dependencies, making them suitable for various applications, from real-time systems to large-scale graph data analysis.


\subsection{Training strategies} 

In the ML paradigm, various approaches are employed to train models, including supervised, semi-supervised, unsupervised, and self-supervised learning. Each of these strategies contributes significantly to the model's performance, generalization capabilities, and scalability. The learning method has a big effect on how well the model fits the data. This is particularly relevant for graph-based learning, where the complexity of the graph structures and the amount of labeled data available change significantly across different domains \cite{atitallah2020leveraging}. 
Graph-based learning introduces unique challenges, as graphs often encode intricate relationships between nodes and edges. The availability of labeled data is usually a challenge, which leads to exploring more flexible learning approaches such as semi-supervised, unsupervised, or self-supervised learning \cite{qiao2018data}. These approaches allow models to learn effectively from both labeled and unlabeled data or even from the structure of the graph itself. In the following sections, we provide a comprehensive review of the learning methods utilized in Graph Mamba models, highlighting how each approach is tailored to address the specific challenges of graph-based learning across various applications.

\subsubsection{Supervised Learning Learning}
Supervised Learning is the most used approach in many graph-based applications, especially in scenarios where high-quality labeled data is available \cite{khoshraftar2024survey}. In this method, models learn to predict the correct output based on the input using an annotated dataset. 
The supervised learning method allows Graph Mamba models to learn directly from labeled data, where the correct outputs are provided. This lets the model make more accurate predictions because it can change its internal representations repeatedly to eliminate mistakes \cite{montagna2024topological}. In addition, supervised learning often leads to faster model convergence because the model knows precisely what output it should achieve at each training step  
\cite{li2024stg}. This lowers the need for extensive fine-tuning, allowing the Graph Mamba model to learn more rapidly from labeled graph data. 

\subsubsection{Semi-Supervised Learning}
In semi-supervised learning, both labeled and unlabeled data are used to train the model \cite{zhu2022introduction}. This method of learning is especially useful in scenarios where labeled data is scarce but unlabeled data is abundant, a common situation in real-world graph datasets \cite{song2022graph}. By combining the advantages of supervised and unsupervised learning, semi-supervised learning enables Graph Mamba models to leverage the structure of the graph and improve predictions even with limited labeled data. 
Semi-supervised learning is especially effective for dynamic or heterogeneous networks where the interconnections among nodes and edges change over time and identifying all data points could be challenging \cite{behrouz2024brain,zhao2024grassnet}. This strategy helps the model learn better generalization by using data from both labeled and unlabeled parts of the graph. This makes it better at predicting results for nodes that aren't labeled.

\subsubsection{Self-Supervised Learning}

Self-supervised learning (SSL) is an emerging paradigm in which models learn from the data's inherent structure. In the context of graph-based learning, SSL has gained significant attention as it allows models to uncover meaningful patterns from the graph’s topology and node features \cite{wu2021self,xie2022self,ben2024strengthening,benjdira2024dm}. To do this, pretext tasks are added to the model to help it understand the data before the main task is conducted. 
In Graph Mamba models, SSL enables the model to capture long-range dependencies within the graph without needing labeled data. This makes it especially useful for dynamic systems, where labeled data is often sparse or unavailable. SSL also enhances the scalability of Graph Mamba models, as they can train on large, real-world graph datasets without being limited by the availability of labels \cite{ding2024combining,tang2023modeling}.

\subsubsection{Unsupervised Learning}
Unsupervised learning is a powerful method for discovering hidden patterns or structures within data without requiring labels \cite{ju2024comprehensive}. In the context of Graph Mamba models, unsupervised learning is utilized to derive significant representations of nodes and edges. These representations can subsequently be employed for downstream tasks, including graph generation or classification.
Following this method, Graph Mamba models are very flexible in areas with less supervision because they can adapt to several applications using just the graph's structure \cite{wang2024graph,li2024state}. The main advantage of this method is the elimination of the requirement for human-annotated data.
 
To conclude, training strategies play a crucial role in the success and adaptability of Graph Mamba models. Supervised learning remains the go-to approach when labeled data is abundant, providing precise and rapid model convergence. Semi-supervised learning bridges the gap between labeled and unlabeled data, particularly effective for real-world scenarios where labeled data is scarce. Self-supervised learning stands out for its ability to leverage the graph’s inherent structure, enabling models to uncover meaningful patterns and scale efficiently to large datasets. Lastly, unsupervised learning empowers Graph Mamba models to discover hidden relationships and structures within graph data. 

\section{Review of Recent Applications of Graph Mamba}
\label{sec4}
By incorporating SSMs, Graph Mamba enables the representation of complex spatial and temporal dependencies in graph structures. These models excel in general-purpose applications like social networks, in addition to other specialized fields such as healthcare, biosignal analysis, and heterogeneous graph learning. This section reviews the diverse applications of Graph Mamba, highlights the innovations introduced in each domain, and examines the critical research publications that showcase its effectiveness. 

\subsection{General-Purpose Graph Learning and Benchmarking}

The convergence of ML with graph data has led to groundbreaking advances across different domains. However, traditional graph learning methods have been limited in their ability to capture long-range dependencies and handle dynamic graph structures efficiently. To address these challenges, SSMs have emerged as a robust framework, enhancing graph learning by integrating temporal modeling and better management of complex relationships within graph data. This method extends the capabilities of graph-based methods by providing a mechanism to capture both spatial and temporal dynamics.

For general purposes and benchmarking, SSMs play an essential role in advancing ML on dynamic graphs by providing a structured way to incorporate long-range dependencies. An example of this advancement is Graph State Space Convolution (GSSC) \cite{huang2024can}, a major step forward in this field. It changes SSMs to work with graph data while keeping permutation invariance and capturing long-range dependencies. This approach primarily focuses on dynamic graphs across various domains, enabling it to handle evolving structures in complex systems. GSSC creates a new convolutional mechanism that uses SSMs in graph settings. This keeps essential graph properties like permutation invariance and models long-range dependencies more efficiently. This method combines global permutation-equivariant set aggregation and factorized graph kernels to quickly and easily figure out the structure of a graph.
The model applies to general graph learning tasks, with notable applications in molecular biology and social networks, where understanding complex, dynamic relationships is essential.

In another prominent contribution, Behrouz \textit{et al.} \cite{behrouz2024graph} introduced a new ML class for graphs utilizing the Selective State Space Model (SSSM). They developed a Graph Mamba Network (GMN) to adapt SSSMs to graph data for heterogeneous and dynamic graphs. The Mamba architecture enables permutation-invariant graph learning with local encoding and token ordering. This approach helps the model to handle different applications like social networks and molecular graphs. Their innovation significantly reduces memory and compute overhead and improves the model's long-range relationships.
In order to capture long-range dependencies, Wang \textit{et al.} \cite{wang2024graph} presented Graph-Mamba, a novel dynamic graph learning technique. Through the integration of a graph-based node prioritization mechanism with SSSMs, Graph-Mamba maximizes long-range sequence modeling while preserving computational efficiency. The model improves on regular GNNs by using Mamba blocks, which let the model choose nodes based on input. This makes it better at finding key connections over long sequences. Graph-Mamba is particularly suited for tasks related to long-range graph sequences, such as traffic prediction and social network analysis, where understanding distant relationships in dynamic systems is necessary.

Some works have addressed temporal challenges in graph learning through the innovative application of SSMs. In \cite{li2024dyg}, Li \textit{et al.} proposed the  DYGMAMBA, a Continuous-Time Dynamic Graph (CTDG) model designed to handle long-term temporal dependencies and efficiently manage large interaction histories within dynamic graphs. The model uses SSM to represent sequences of historical node interactions while preserving essential temporal details for accurate future predictions. The model was tested on dynamic link prediction tasks, with results showing it achieves state-of-the-art performance on several CTDG datasets. Besides, Li \textit{et al.} \cite{li2024state} proposed the GRAPHSSM model, an SSM designed specifically for temporal graphs to address challenges in modeling time-evolving structures. It integrates the GHIPPO framework to extend SSM theory to graphs, using Laplacian regularization for capturing evolving topological structures. GRAPHSSM also includes a mixed discretization approach to handle unobserved graph mutations efficiently.

Table \ref{tab:3} illustrate a summary about these works. These advancements underscore the utility of SSMs in enhancing general-purpose graph learning, particularly for dynamic graph structures. By addressing limitations in conventional methods, such as limited capacity for long-range dependencies and dynamic structure adaptation, SSM-based models like GSSC, GMN, Graph-Mamba, DYGMAMBA, and GRAPHSSM push the boundaries of graph ML applications, providing robust solutions that scale effectively across diverse domains.
\begin{table}[!htbp]
\centering
\caption{Overview of Approaches in Graph Mamba Works: General-Purpose Graph Learning and Benchmarking}
\label{tab:3}
\resizebox{\textwidth}{!}{
\begin{tabular}{p{1cm}p{2cm}p{2cm}p{2cm}p{4cm}p{4cm}}
\hline
\textbf{Work} & \textbf{Graph Type} & \textbf{Architecture} & \textbf{Application Domain} & \textbf{Contributions} & \textbf{Limitations} \\
\hline
 Huang \textit{et al.} (2024) \cite{huang2024can} & Dynamic graphs & Graph State Space Convolution (GSSC), Long-range dependencies & Molecular biology, Social networks & Introduces GSSC, combining SSM with graph learning; efficiently captures long-range dependencies while maintaining permutation invariance & Limited scalability with extremely large datasets; fails with very high-dimensional graph applications \\
\hline 
Behrouz \textit{et al.} (2024) \cite{behrouz2024graph} & Heterogeneous and dynamic graphs & SSSM, Local encoding, Token ordering & Social networks, Molecular graphs & Adapts SSSMs for dynamic and heterogeneous graph data; reduces memory and compute overhead & Integration of SSMs into large-scale graph networks is still complex; scalability and efficiency need improvements \\
\hline 
Wang \textit{et al.} (2024) \cite{wang2024graph} & Dynamic graphs & Node prioritization mechanism, SSSM for long-range sequence modeling & Traffic prediction, Social networks & Enhances sequence modeling through node prioritization; excels in capturing long-range dependencies in dynamic graphs & Less effective for non-sequential graph data; focus on long-range dependencies \\
\hline 
Li \textit{et al.} (2024) \cite{li2024dyg} & Continuous-Time Dynamic Graph (CTDG) & Node-Level SSM, Time-Level SSM)	& Dynamic link prediction	&Novel two-level SSM for handling long-term dependencies, dynamic selection of relevant temporal information, & Primarily evaluated on link prediction; limited testing on other CTDG tasks and interaction frequency variations in real-world networks \\
\hline 
Li \textit{et al.} (2024) \cite{li2024state} & Temporal Graph	& GHIPPO-based SSM with mixed discretization	& Node classification on evolving graphs & Introduces GHIPPO for memory compression on dynamic graphs, mixed discretization for handling mutations, efficient handling of long temporal sequences &	Focused on discrete-time temporal graphs, limited exploration of varying node sets and continuous dynamics\\
\hline
\end{tabular}}
\end{table}
\subsection{Knowledge and Heterogeneous Graphs}

The development of ML models for heterogeneous and knowledge graphs has opened new avenues for handling multi-relational data structures. These graphs are commonly used in fields such as social networks, recommendation systems, and knowledge representation, where diverse entities and connections need to be effectively modeled. SSSMs offer a solid way to improve graph learning by including temporal dynamics and accurately capturing the complex relationships that exist in different types of graphs.

To address the challenges posed by heterogeneous graphs, Pang \textit{et al.} \cite{pan2024hetegraph} developed HeteGraph-Mamba, a model that uses SSMs to manage varied data structures with multiple node and edge types. It records different sorts of nodes and their connections using different edge types. The model includes a graph-to-sequence conversion approach that reduces complicated graph data while keeping structural subtleties. This strategy is especially useful in applications such as knowledge graphs and heterogeneous networks, as illustrated by the IMDB dataset, in which entities (e.g., films, actors, and directors) are linked together via multiple connections.

Montagna \textit{et al.} \cite{montagna2024topological} designed a novel architecture that combines SSM with topological DL for analyzing higher-order structures, specifically simplicial complexes. The model constructs sequences for each node based on neighboring cells of various ranks and processes these sequences through Mamba blocks to enable information propagation across all ranks.

To address limitations in spectral GNNs, Zho \textit{et al.} \cite{zhao2024grassnet} introduced GrassNet, the first model to apply SSMs for spectral filtering. 
This model treats the graph spectrum as a sequence and enhances expressivity by capturing correlations across spectral components.

SSMs have shown remarkable potential in enhancing ML models for knowledge and heterogeneous graphs. By addressing challenges in modeling complex relationships and long-range dependencies, SSMs enable efficient and scalable graph learning. Approaches, like HeteGraph-Mamba and GrassNet, highlight how Graph Mamba can improve computational efficiency, scalability, and the ability to manage evolving graph structures.

\begin{table}[!htbp]
\centering
\caption{Overview of Approaches in Graph Mamba Works: Knowledge and Heterogeneous Graphs}
\label{tab:4}
\resizebox{\textwidth}{!}{
\begin{tabular}{p{1cm}p{2cm}p{2cm}p{2cm}p{4cm}p{4cm}}
\hline
\textbf{Work} & \textbf{Graph Type} & \textbf{Architecture} & \textbf{Application Domain} & \textbf{Contributions} & \textbf{Limitations} \\
\hline
Pan \textit{et al.} (2024) \cite{pan2024hetegraph} & Heterogeneous graphs & SSSM with graph-to-sequence conversion & Knowledge graphs, Heterogeneous networks & Introduces graph-to-sequence conversion; captures structural subtleties and reduces graph complexity & Challenges with extremely large heterogeneous graphs with sparse connections \\
\hline

Montagna \textit{et al.} (2024) \cite{montagna2024topological} & simplicial complexes &	SSM with Mamba Blocks, sequence-based processing	& Classification and Regression & Introduces SSM for higher-order interactions, efficient memory-aware batching, & Evaluated on static datasets; limited interpretability and complexity \\
\hline  

Zhao \textit{et al.} (2024) \cite{zhao2024grassnet} & Spectral Graph &	SSM-based spectral filters with sequential spectrum processing	& Node classification (citation networks, web graphs, co-purchase networks) & SSM-based spectral filtering with sequence handling, enhanced robustness to spectral noise and perturbations, superior accuracy and scalability & Primarily evaluated on semi-supervised node classification; complex design may require tuning for different graph structures  \\
\hline
\end{tabular}}
\end{table}

In conclusion, the reviewed approaches, as summarised in Table \ref{tab:4}, highlight SSMs' significant advancements in graph Learning, particularly in dynamic and heterogeneous environments. These models address the limitations of traditional graph-based ML methods and enable more efficient handling of long-range relationships and evolving structures. Innovations such as GSSC demonstrate the role of Graph Mamba in enhancing scalability, improving computational efficiency, and maintaining essential graph properties. 

\subsection{Traffic and Environmental Forecasting}

Traffic and environmental forecasting pose significant challenges that demand accurate modeling of complex spatiotemporal dynamics \cite{tedjopurnomo2020survey}. These fields require systems that can capture evolving patterns, such as fluctuating traffic flows, weather changes, and environmental shifts, across vast networks of interconnected nodes. Traditional methods often fall short in handling long-range dependencies and non-linear relationships inherent in these dynamic datasets \cite{shaygan2022traffic}. Graph Mamba-based models, integrated with SSMs, have recently emerged as powerful solutions, offering enhanced capability to model both spatial structures and temporal sequences.

For example, Li \textit{et al.} \cite{li2024stg} introduced a Spatial-Temporal Graph Mamba (STG-Mamba) approach, which leverages SSSM for STG learning. The proposed approach addresses challenges in forecasting dynamic STG data such as traffic and weather networks, and allows more efficient handling of spatial-temporal data through adaptive feature selection. The STG-Mamba model is tailored for spatial-temporal forecasting and is useful in domains such as traffic prediction, urban flow forecasting, and environmental data analysis.

In \cite{choi2024spot}, Choi \textit{et al.} developed SpoT-Mamba, a spatio-temporal graph (STG) forecasting framework to address complex long-range dependencies in spatial-temporal data. 
The architecture consists of multi-way walk sequences (depth-first search, breadth-first search, and random walks) that extract node-specific structural information. Mamba blocks process these sequences to embed the spatial structure of each node's neighborhood and capture dependencies over lengthy temporal epochs. The model’s temporal scanning with Mamba-based blocks allows it to selectively retain or ignore information over time, efficiently capturing repeating patterns in the data. 

In \cite{zhang2024enhanced}, Zhang \textit{et al.} developed an approach for predicting trajectories in multi-agent systems, such as autonomous vehicles and pedestrians. This model integrates GNNs with a specialized SSM, which enables accurate control variable inference for each agent. By modeling control states with physical significance, the approach achieves high predictive accuracy and interpretability. Evaluations across several datasets show that this method outperforms benchmarks, underscoring its potential to improve safety and efficiency in autonomous systems. 

In \cite{zhou2024graph}, Zhou \textit{el al.} proposed a Graph Convolution Network -based State Space Model (GSSM) for predicting wireless traffic from call detail records (CDRs). They leveraged both fixed and adaptive weighted matrices to capture intricate spatial dependencies among base stations. The GCN is employed to accurately estimate SSM parameters. Furthermore, a mixture of Gaussian hypothesis is incorporated into the SSM to enhance flexibility in traffic prediction.

In summary, Graph Mamba models with SSM integration represent a significant jump forward in traffic and environmental forecasting. These models provide a flexible framework for capturing spatial-temporal dependencies and enabling more accurate predictions across diverse conditions. Despite their strengths, computational demands and scalability remain areas for improvement, especially for applications requiring real-time processing over extensive networks, as depicted in Table \ref{tab:5}. 

\begin{table}[!htbp]
\centering
\caption{Overview of Approaches in Graph Mamba Works: Traffic and Environmental Forecasting}
\label{tab:5}
\resizebox{\textwidth}{!}{
\begin{tabular}{p{1cm}p{2cm}p{2cm}p{2cm}p{4cm}p{4cm}}
\hline
\textbf{Work} & \textbf{Graph Type} & \textbf{Architecture} & \textbf{Application Domain} & \textbf{Contributions} & \textbf{Limitations} \\
\hline
 Li \textit{et al.} (2024) \cite{li2024stg}  & STG &	Graph Selective State Space Blocks (GS3B) with KFGN and ST-S3M &	Traffic prediction, urban flow forecasting, environmental data analysis	& First adaptation of SSSM for STG, Kalman Filtering-based dynamic updates for temporal granularity, Reduced computational costs &	Limited generalizability tested across only three datasets; model interpretability is challenging; performance may degrade with incomplete/noisy data  \\
\hline 
Choi \textit{et al.} (2024) \cite{choi2024spot} & STG & Multi-way Walk Sequences, Mamba-based Temporal Scans	& Traffic forecasting & Node-specific multi-way walks for diverse local and global information, selective state-space mechanism reducing computational cost and improving efficiency &	Limited real-world datasets tested; potential challenges in scalability with larger networks; model interpretability remains challenging\\
\hline 
Zhang \textit{et al.} (2024) \cite{zhang2024enhanced}& Multi-Agent Interaction Graph  & Mixed Mamba with Control Variable Inference, GNN integration  & Autonomous Systems, Multi-Agent Trajectory Prediction  & Accurately infers control variables for each agent, high interpretability and predictive accuracy  &  Scalability to real-time applications could be improved, applicability may vary depending on agent behavior complexity\\ \hline  
Zhou \textit{et al.} (2024) \cite{zhou2024graph} & STG & Graph convolution network based state space model & Traffic data prediction & Adaptively learns the graph structure of base stations and effectively captures spatio-temporal traffic patterns & Challenges in large-scale, complex network traffic prediction remain unsolved \\ \hline 
Pan \textit{et al.} (2024) \cite{pan2024hetegraph} & Heterogeneous graphs & SSSM with graph-to-sequence conversion & Knowledge graphs, Heterogeneous networks & Introduces graph-to-sequence conversion; captures structural subtleties and reduces graph complexity & Challenges with extremely large heterogeneous graphs with sparse connections \\

\hline
\end{tabular}}
\end{table}

\subsection{Healthcare and Biosignals}
Healthcare and biosignal analysis has recently emerged as critical areas for applying advanced ML models, especially those using graph-based approaches \cite{li2022graph}. The complex and dynamic characteristics of biological and medical data, including electroencephalograms (EEG), functional MRI (fMRI), and patient medical records, require models that can effectively capture spatial and temporal dependencies. GNNs combined with SSMs have emerged as powerful tools, providing a robust framework for learning from both structured and time-evolving data. 
Graph-based methods are well-suited for healthcare applications because they can model the intricate relationships between various biological components (e.g., neurons in the brain, organs in the body) \cite{paul2024systematic}. By incorporating state-space models, these methods get the ability to capture temporal patterns. Such hybrid approaches are practical in seizure detection, sleep staging, and disease diagnosis, where both spatial and temporal relationships play a vital role. In this section, we review several notable works that leverage Graph Mamba models to address challenges in healthcare and biosignal analysis.

For instance, Tang \textit{et al.} \cite{tang2023modeling} proposed the GraphS4mer for modeling spatiotemporal dependencies in biosignals through the use of dynamic graphs. The architecture integrates stacked S4 layers to capture long-range temporal dependencies, employs Graph Structure Learning (GSL) for the dynamic learning of graph structures over time, and utilizes Graph Isomorphism Networks (GIN) to model spatial dependencies. GraphS4mer exhibits enhanced performance in medical applications, including seizure detection, sleep staging, and ECG classification. For the same objective, BrainMamba \cite{behrouz2024brain}, a sophisticated model, has been developed for brain activity encoding through SSSMs. This work presents two essential modules: Brain Network Mamba (BNMamba) for graph learning and Brain Timeseries Mamba (BTMamba) for encoding multivariate brain signals. The model integrates spatial brain network data with temporal time series data to capture long-range dependencies in multivariate brain signals effectively. BNMamba employs a graph learning methodology utilizing Message-Passing Neural Networks (MPNN) to encode local interdependencies among brain units. The process encompasses tokenization for nodes (brain units) and selectively models long-range dependencies through SSSMs. On the other hand, BTMamba encodes time-varying brain signals by integrating an SSM for each time series with a Multi-Layer Perceptron (MLP) to fuse information across variables. This model outperforms state-of-the-art methods across seven datasets, including fMRI, EEG, and MEG, for tasks like ADHD classification and seizure detection.

In the domain of patient data, Xu \textit{et al.} \cite{xu2024identifying} introduced a dynamic graph-based method for identifying patient subphenotypes related to Sepsis and Acute Kidney Injury (SAKI). The model utilizes a dynamic graph structure that adjusts according to patient data, illustrating variations in the health of critically ill patients over time. This approach incorporates Multimodal Feature Fusion, Adaptive Latent Graph Inference, and SSMs to facilitate the dynamic learning of graph structures from multimodal data, including demographics, diagnostics, lab results, and vital signs. The GNN framework facilitates local neighborhood encoding through message-passing and allows the model to learn the relationships among various features. 

In addition to biosignals, the integration of graph-based methods with SSSMs has enhanced the modeling of metabolic pathways. For example, in \cite{aghaee2024graph}, a GNN representation with SSM method for metabolic pathways is proposed. This approach leverages SSSMs to generate dynamic metabolic graphs that represent biological pathways and the temporal changes in metabolites and reactions. The Metabolic Graph Neural Network (MGNN) uses SSSMs to model how these processes change over time and accurately capture metabolic activity that is always changing. It is applied in modeling biological and chemical processes, such as oxidative stress pathways in bacteria.
In \cite{ding2024combining}, Ding \textit{et al.} present a novel pipeline for predicting progression-free survival in early-stage lung adenocarcinoma (LUAD) that combines message-passing GNNs and state space modeling. A GAT \cite{velivckovic2017graph} served as the GNN component, while Mamba was used as the SSM to capture local and global tissue interactions within Whole Slide Images (WSIs), modeled as graphs. The authors also investigated the impact of different node feature types and tile sampling strategies on model performance.

Moreover, graph-based models integrated with state-space models (SSMs) have notably advanced healthcare image processing \cite{jiao2022graph}. By representing image elements, such as anatomical structures, as nodes and their relationships as edges, these models are able to learn both local and global interactions \cite{szeliski2022computer, chen2024survey}. The addition of SSMs strengthens these models by managing time-varying and sequential data. This is especially beneficial for dynamic imaging modalities, such as MRI sequences or real-time ultrasound \cite{rahman2024mamba}.
For example, Zhang \textit{et al.} \cite{zhang2024gm} introduced a hybrid method integrating GNNs with the UNet architecture to enhance medical image segmentation. The model uses a Graph Mamba framework and a dynamic graph structure to depict the relationships between image features. This allows the segmentation process to adapt to both local and global dependencies in the data. The Parallel Channel Spatial Attention (PCSA) and Graph Cross Attention (GCA) modules are the significant components that enhance the model's ability to focus on long-range dependencies and important features. The model demonstrates superior performance compared to traditional methods in tasks such as skin lesion segmentation and organ segmentation, achieving high accuracy on datasets including ISIC2017, ISIC2018, and Synapse. In addition, Zhou \textit{et al.} \cite{zhou2024efficient} introduced the Gender-adaptive Graph Vision Mamba (GGVMamba) framework to address pediatric bone age assessment (BAA) using raw X-ray images. GGVMamba leverages a novel directed scan module and a graph Mamba encoder for efficient feature extraction and enhanced context integration. The directed scan module captures the local context in multiple directions, while the graph Mamba encoder uses a bidirectional compression model to robustly learn inter-region relationships. Additionally, a gender-adaptive strategy dynamically selects gender-specific graph structures to reduce inter-gender discrepancies. The model performs well across many datasets with excellent gender constancy and low GPU use.

Based on RGB images, Dong \textit{et al.} \cite{dong2024hamba} introduced a framework for robust 3D hand reconstruction. It addresses challenges like articulated motion, self-occlusion, and object interactions by combining graph learning with SSM. The main contribution is the Graph-guided Bidirectional Scan (GBS), which learns joint relations and spatial sequences, using 88.5\% fewer tokens compared to attention-based methods. The model demonstrates superior performance compared to current state-of-the-art methods in benchmarks such as FreiHAND and HO3Dv2 for 3D hand reconstruction. The framework is applied in 3D hand reconstruction within computer vision, with applications in fields such as robotics, animation, human-computer interaction, and AR/VR.

In summary, the combination of structured SSMs with graph-based techniques offers an effective method for analyzing data that changes over time in healthcare applications. From biosignal processing to metabolic pathway modeling and patient health prediction, these methods provide promising avenues for improving diagnostic accuracy, personalized treatment, and real-time monitoring. However, there are still challenges, particularly regarding scalability when applied to large datasets and the reliance on predefined network structures. Additionally, the models' computational demands, especially in real-time applications like continuous patient monitoring, require further optimization. Future research should focus on refining these models to enhance their adaptability and efficiency. Moreover, efforts to integrate unsupervised or semi-supervised learning approaches could further reduce the need for manually curated datasets, making these models more robust in clinical settings with limited labeled data. Table \ref{tab:6} provides an overview of Graph Mamba approaches proposed to solve challenges in healthcare and biosignals.

\begin{table}[ht]
\centering
\caption{Overview of Approaches in Graph Mamba Works: Healthcare and Biosignals}
\label{tab:6}
\resizebox{\textwidth}{!}{
\begin{tabular}{p{1cm}p{2cm}p{2cm}p{2cm}p{4cm}p{4cm}}
\hline
\textbf{Work} & \textbf{Graph Type} & \textbf{Architecture} & \textbf{Application Domain} & \textbf{Contributions} & \textbf{Limitations} \\
\hline
Tang \textit{et al.} (2023) \cite{tang2023modeling} & Dynamic Graph & S4 layers for temporal dependencies, GSL, GIN & Seizure detection, Sleep staging, ECG classification & Integrates long-range temporal dependencies with spatial graph learning, dynamic structure learning & Scalability in larger, more complex biosignal datasets could be explored further \\
\hline
Behrouz \textit{et al.} (2024) \cite{behrouz2024brain} & Brain Network Graph & BNMamba for graph learning, BTMamba for time series & fMRI, EEG, MEG, Seizure detection, ADHD classification & Combines brain network data with time-series for long-range dependencies & High computational complexity in real-time settings \\
\hline
Xu \textit{et al.} (2024) \cite{xu2024identifying} & Dynamic Health Graph & Multi-modal feature fusion & Sepsis and Acute Kidney Injury subphenotypes & Learns dynamic patient data relationships, integrates multi-modal medical data & Applicability in non-critical conditions remains untested \\
\hline
Aghaee \textit{et al.} (2024) \cite{aghaee2024graph} & Metabolic Pathway Graph & SSSMs for dynamic metabolic graphs & Metabolic pathways, Biological process modeling & Captures time-varying metabolic activities, handles oxidative stress pathways & Dependence on known metabolic networks limits applicability to unknown structures \\
\hline
Ding \textit{et al.} (2024) \cite{ding2024combining} & Whole Slide Image Graph & GAT, Mamba for SSM & Early-stage lung adenocarcinoma, Progression prediction & Captures local and global tissue interactions from WSIs, explores node feature types & Model refinement needed for wider clinical integration, further validation required \\
\hline
Dong \textit{et al.} (2024) \cite{dong2024hamba} & Dynamic Spatial Graph   &  - Graph-guided Bidirectional Scan  - Combines SSM with graph learning  &  3D hand reconstruction &   - Uses 88.5\% fewer tokens - Excels in articulated motion and  self-occlusion tasks    &   - Focuses on 3D hand reconstruction -Scalability to other 3D objects is untested   \\
\hline
Zhang \textit{et al.} (2024) \cite{zhang2024gm} &   Dynamic Feature-based Graph  &   - Hybrid GNN-UNet architecture - PCSA and GCA modules   &  Medical image segmentation  &   - Captures long-range dependencies - High accuracy in skin lesion and organ segmentation  &   - Computationally intensive for real-time applications   \\
\hline
Zhou \textit{et al.} (2024) \cite{zhou2024efficient} & Pediatric bone structure graph & Directed Scan Module with Graph Mamba Encoder; gender-adaptive graph structure & Pediatric bone age assessment from X-rays & Efficiently processes X-ray images for age prediction; adapts model to gender-specific bone growth patterns & Performance may vary across diverse populations; limited testing in non-pediatric age groups \\
\hline
\end{tabular} 
}
\end{table}
\raggedbottom

\subsection{Remote Sensing}

Remote sensing data often involves high-dimensional, multi-spectral, or temporal information captured over large geographic areas, making it complex to analyze. Applications such as land cover classification, environmental monitoring, and urban planning rely on accurate modeling of both spatial and spectral dependencies across varied data types, including hyperspectral images, satellite imagery, and LiDAR scans. Traditional methods face challenges in capturing these multi-scale dependencies and handling the extensive data volume. Graph Mamba models have emerged as powerful tools to tackle these challenges \cite{khlifi2023graph}.
For instance, Yang \textit{et al.} \cite{yang2024graphmamba} introduced GraphMamba, an innovative approach designed for efficient hyperspectral image (HSI) classification. The model uses both HyperMamba for processing spectral data and SpatialGCN for extracting spatial features. This makes the architecture very efficient at dealing with the unique properties of hyperspectral data. Table \ref{tab:7} provides an overview of this work and highlights its main contributions and limitations.

Graph Mamba models, with their ability to integrate spatial and spectral processing seamlessly, represent a significant advancement in remote sensing applications. By addressing challenges such as high-dimensionality and multi-scale dependencies, they demonstrate the potential to revolutionize tasks like land cover classification and environmental monitoring. However, challenges remain in scaling these models for very large datasets and ensuring generalizability across varied remote sensing contexts. The integration of Graph Mamba with remote sensing datasets opens new avenues for research, combining cutting-edge graph-based methodologies with pressing environmental and geographic challenges.
\begin{table}[!htbp]
\centering
\caption{Overview of Approaches in Graph Mamba Works: Remote Sensing }
\label{tab:7}
\resizebox{\textwidth}{!}{
\begin{tabular}{p{1cm}p{2cm}p{2.5cm}p{2cm}p{4.2cm}p{3cm}}
\hline
\textbf{Work} & \textbf{Graph Type} & \textbf{Architecture} & \textbf{Application Domain} & \textbf{Contributions} & \textbf{Limitations} \\
\hline
 Yang \textit{et al.} (2024) \cite{yang2024graphmamba}   &   Static Spatial, Spectral Graph  & HyperMamba and SpatialGCN, SSSMs for high-dimensional data   &  Hyperspectral image classification, Remote sensing  &  Efficiently handles hyper-spectral data, Captures long-range spatial dependencies &  Generalizability to non-hyperspectral data is limited   \\
\hline
\end{tabular}}
\end{table}

\subsection{Financial Markets and Economic Forecasting}

Financial markets present a unique set of challenges for predictive modeling, including high volatility, long-range temporal dependencies, and the need for near-instantaneous data processing \cite{sonkavde2023forecasting}. In traditional forecasting models, a set of limitations exists, including the capture of intricate relationships and dependencies among financial variables over extended time horizons. Graph Mamba-based models, offer a powerful solution by enabling detailed temporal dependency modeling and efficient handling of high-dimensional data \cite{ashtiani2023news}. 
In Table \ref{tab:8}, the SAMBA model, introduced by Mehrabian \textit{et al.} \cite{mehrabian2024mamba}, presents an advanced framework for stock return forecasting by combining the Mamba architecture with GNNs. This integration allows SAMBA to capture complex and long-range dependencies in financial data while maintaining computational efficiency. The model's core components are the Bidirectional Mamba (BI-Mamba) block and Adaptive Graph Convolution (AGC) block. These two blocks work together to efficiently process long-sequence data. They enable SAMBA to achieve near-linear computational complexity which make it well-suited for real-time trading applications that require both speed and accuracy.
\begin{table}[!htbp]
\centering
\caption{Overview of Approaches in Graph Mamba Works: Financial Markets and Economic Forecasting}
\label{tab:8}
\resizebox{\textwidth}{!}{
\begin{tabular}{p{1cm}p{2cm}p{2cm}p{2.5cm}p{4cm}p{3cm}}
\hline
\textbf{Work} & \textbf{Graph Type} & \textbf{Architecture} & \textbf{\makecell{Application\\ Domain}} & \textbf{Contributions} & \textbf{Limitations} \\
\hline
 
\makecell{Mehra-\\bian} \textit{et al.} (2024) \cite{mehrabian2024mamba} & Financial interaction graph & BI-Mamba Block with AGC Block & Stock return forecasting, high-frequency trading, portfolio management & Efficiently captures long-range dependencies in financial data with near-linear computational complexity; real-time adaptability in trading environments & May require further optimization for highly volatile markets \\ 
\hline
\end{tabular}}
\end{table}
The integration of Graph Mamba-based models, such as SAMBA, into financial market forecasting represents a significant contribution for addressing the challenges posed by high-dimensional data. However, challenges remain in scaling these models for larger datasets and adapting them to diverse financial instruments and markets. Future research could focus on enhancing scalability, improving interpretability for regulatory compliance, and extending these models to multi-modal data integration. 

\subsection{Sentiment Analysis}
Sentiment analysis is defined as the process of text analysis to determine if the emotional tone of the message is positive, negative, or neutral. In \cite{lawan2024mambaforgcn}, Lawan \textit{el al.} proposed Aspect-Based Sentiment Analysis (ABSA) that goes beyond overall sentiment analysis by identifying and evaluating sentiments towards specific aspects of entities within text. Attention mechanisms and neural network models are commonly utilized in ABSA, but syntactic limitations and quadratic complexity prevent them from capturing long-range aspect-opinion word relationships. To overcome these limitations, the authors introduce MambaForGCN, a novel approach that effectively combines syntactic and semantic information. Their model leverages syntax-based Graph Convolutional Network (SynGCN) and MambaFormer modules to capture both short and long-range dependencies. The Multihead Attention (MHA) and Mamba blocks in the MambaFormer module serve as channels to enhance the model's ability to capture these dependencies. Additionally, they employed the Kolmogorov-Arnold Networks (KANs) \cite{liu2024kan} gated fusion to adaptively integrate the representations from both modules. Experimental results on three benchmark datasets demonstrated the effectiveness of their approach, outperforming state-of-the-art baselines. An overview about this work is presented in Table \ref{tab:9}.
\begin{table}[!htbp]
\centering
\caption{Overview of Approaches in Graph Mamba Works: Sentiment Analysis}
\label{tab:9}
\resizebox{\textwidth}{!}{
\begin{tabular}{p{1cm}p{1.8cm}p{2cm}p{2cm}p{3cm}p{6cm}}
\hline
\textbf{Work} & \textbf{Graph Type} & \textbf{Architecture} & \textbf{Application Domain} & \textbf{Contributions} & \textbf{Limitations} \\
\hline
  
Lawan \textit{el al.} (2024) \cite{lawan2024mambaforgcn} & Syntactic graph & MambaForGCN & Aspect-based sentiment analysis & Investigates the potential of combining Mamba, Transformer, and GCN models to improve ABSA performance &  The integration of multiple complex modules, such as SynGCN and MambaFormer, can increase the computational cost, especially for large-scale datasets; it may still struggle to fully capture complex contextual nuances in ambiguous or sarcastic language \\
\hline
\end{tabular}}
\end{table}
The integration of graph-based models and state-space mechanisms demonstrates significant potential for addressing the limitations of traditional methods in complex analysis tasks. By combining structural and contextual modeling, such approaches enable a deeper understanding of intricate relationships within data. Future efforts could focus on applying these models to other datasets, multilingual scenarios, and real-time analysis. This could lead to interesting contributions in areas like as social media monitoring, evaluating client feedback, and studying market trends.

\section{Performance Evaluation and Benchmarking Datasets}
\label{sec5}
This section focuses on the key evaluation metrics particularly used for Graph Mamba. In addition to metrics, we explore the data sources that are used with Graph Mamba methodologies. 

\subsection{Evaluation Metrics and Validation}
Evaluation is critical for guiding the development of Graph Mamba architectures. Metrics can be categorized into several groups based on their focus, such as training performance, clustering quality, prediction accuracy, and regression performance. This section provides a structured overview of these metrics, specifically tailored to Graph Mamba, while excluding task- or domain-specific measures.

\subsubsection{Prediction Performance Metrics}

For classification tasks, this group of metrics evaluates the model’s ability to accurately predict outcomes. These measures are essential for understanding the overall performance and reliability of predictions.

\begin{itemize}
    \item \textbf{Accuracy:}
This metric, often referred to as overall accuracy, assesses the model's ability to correctly classify both positive and negative instances. It's calculated by dividing the sum of true positives (TP) and true negatives (TN) by the total number of samples which includes the false positives (FP) and the false negatives (FN).
\begin{equation}
\text { Accuracy }=\frac{T P+T N}{T P+T N+F P+F N}
\end{equation}

    \item \textbf{Precision:}
Precision measures how accurate a model's positive predictions are. It calculates the proportion of correct positive predictions out of all positive predictions made.
A high precision indicates that when the model predicts a positive class, it is likely to be correct.
\begin{equation}
\text { Precision }=\frac{T P}{T P+F P}
\end{equation}

    \item \textbf{Recall:}
Recall, or sensitivity, measures how well a model identifies all positive instances. It calculates the proportion of actual positive instances that the model correctly identifies as positive.
A high recall indicates that the model is good at finding all positive instances, even if it might also incorrectly identify some negative instances as positive.
\begin{equation}
\text { Recall }=\frac{T P}{T P+F N}
\end{equation}


    \item \textbf{F1-score:}
The F1 Score is a metric that assesses a model's performance by considering both precision and recall. It's calculated as the harmonic mean of these two metrics, penalizing extreme values in either. This makes it a valuable tool for situations where a balance between high precision and high recall is desired.
\begin{equation}
\text { F1 }= \frac{2* Precision * Recall}{Precision + Recall} = \frac{2* T P}{2* T P+F P+F N}
\end{equation}

    \item \textbf{Kappa coefficient:}
Cohen's kappa coefficient is a statistical measure used to assess the agreement between two raters who classify a set of items into a fixed number of categories. It measures the level of agreement between classifications generated by an approach and those made by human raters. Values between 0.6 and 0.8 suggest a substantial level of consistency between the two.
\end{itemize}

\subsubsection{Regression Performance Metrics}
Regression evaluation metrics are crucial for assessing the performance of a regression model. They provide a quantitative measure of how well the model predicts continuous numerical values. The choice of the following metrics depends on the specific use case and the desired outcome. 

\begin{itemize}
    \item \textbf{Mean Absolute Error:}
The Mean Absolute Error (MAE) is a widely adopted metric for evaluating the accuracy of continuous variables, such as predicted user ratings in recommendation systems. It quantifies the average absolute deviation between predicted and actual values. The mathematical formulation of MAE is as follows.
\begin{equation}
 \text {MAE} =  \frac{1}{n}\sum_{i=1}^{D}|x_i-y_i|
\end{equation}
where $x$ is the prediction rating and $y$  is the true rating in testing data set, and $n$ is the number of errors.

    \item \textbf{Root mean square error:}
The root mean square error (RMSE) is a commonly used metric to measure the difference between predicted values and actual values. It provides a measure of how spread out these residuals are. It's widely used in regression analysis, forecasting, and ML to evaluate model performance.

 \item \textbf{Modified squared correlation coefficient}
The modified squared correlation coefficient $r_m^2$ is a metric used to assess the generalization performance of the model.
\begin{equation}
r_m^2=r^2\left(1-\sqrt{r^2-r_0^2}\right)
\end{equation}
where $r^2$ denotes the squared correlation coefficient between the model's output and actual scores, while $r_0^2$ represents the squared correlation coefficient with a zero intercept. For a model to be considered adequate, $r_m^2$ should surpass 0.5.

    \item \textbf{Mean Absolute Percentage Error:}
The Mean Absolute Percentage Error (MAPE) is similar to MAE but normalizes the absolute error of each prediction by its corresponding actual value, expressing the result as a percentage. It is a statistical metric used to assess the accuracy of forecasting models. It quantifies the average percentage deviation of predicted values from their corresponding actual values. It is defined by the formula:
\begin{equation}
\text { MAPE }=100 \frac{1}{n} \sum_{t=1}^n\left|\frac{A_t-F_t}{A_t}\right|
\end{equation}
where $A_t$ is the actual value, $F_t$ is the forecast value, and $n$ is the number of fitted points.

    \item \textbf{Negative Log-Likelihood:}
The Negative Log-Likelihood measures how well a model's predicted probabilities match the actual data distribution.

The Average Negative Log-Likelihood (ANLL) and the Final Negative Log-Likelihood (FNLL) are used in the literature to evaluate the graph-Mamba based techniques.
   
\end{itemize}

\subsubsection{Segmentation Metrics}
Segmentation evaluation metrics are crucial in assessing the accuracy and quality of image segmentation models to increase research reliability and reproducibility in the field. They provide quantitative measures to evaluate how well a model can accurately identify and delineate objects or regions of interest within an image. The mean Intersection over Union (mIoU) and the Dice Similarity Coefficient (DSC) are the most commonly used metrics for segmentation. They both range from 0 to 1 where 1 signifies a great similarity between the truth and the prediction.

\begin{itemize}
    \item \textbf{Mean Intersection over Union:}
A widely used performance metric defined by the overlap between the predicted segmentation and the ground truth, divided by the total area covered by the union of the two. It can be calculated for each class independently or for the overall dataset. The metric ranges from 0 to 1, with 1 indicating a perfect match between the predicted and ground truth regions, and 0 indicating no overlap.
    \item \textbf{Dice Similarity Coefficient:}
A statistical measure, also known as the Sørensen–Dice index or simply dice coefficient, is used to evaluate the similarity between two binary segmentation. It is calculated as twice the intersection of the two sets divided by the sum of their individual sizes. The DSC ranges from 0 to 1, with higher values indicating greater similarity.
\end{itemize}

\subsubsection{Clustering Quality Metrics}
This subsection will delve into the evaluation metrics employed for clustering algorithms. These metrics are indispensable for assessing the efficacy of ML algorithms designed to group similar data points. Prior Graph-Mamba research has utilized the Davies–Bouldin Index (DBI) and Calinski–Harabasz Index (CHI) metrics for evaluation. 
\begin{itemize}
    \item \textbf{Davies–Bouldin Index:} A measure of the average similarity of each cluster with its most similar cluster.
    \item \textbf{Calinski–Harabasz Index:} Assesses cluster separation by comparing within-cluster and between-cluster dispersion. Higher values suggest more distinct clusters.
\end{itemize}

\subsubsection{Training Performance Metrics}
To quantify the efficiency and effectiveness of an AI model, a key factor to consider is the training time.

\begin{itemize}
    \item \textbf{Training Time:} This metric measures the duration required for a model to learn patterns from the dataset and optimize its internal parameters. It provides a direct indicator of the computational efficiency of the learning process.
\end{itemize}

\subsection{Benchmarking Datasets and Applications}
Recently, graph Mamba methodologies have gained significant traction in many fields, especially after the introduction of Visual Mamba. This subsection highlights detailed examples of their applications, organized by usability and accompanied by information on dataset utilization and code availability, as summarized in Table \ref{table:data1}.

Additionally, in Table \ref{table:datasets1}, we present a list of some datasets used to evaluate the performance of Graph-Mamba techniques. Detailed metadata and direct links are provided to streamline data access and utilization for research purposes. Many of these datasets were not originally set up as graphs. They include different types of data, like pictures, price data, and text. However, experts have used a variety of methods to turn these datasets into graphs.
This transformation allows the exploration of complex relationships and patterns within the data.
Examples of different types of data transformation into graphs include:

\begin{enumerate}
    \item \textbf{Image Data:} 
\begin{itemize}
    
    \item Superpixel Graphs: Images can be segmented into superpixels, where each superpixel is treated as a node, and edges are established based on adjacency or similarity in color, texture, or intensity. For instance, in \cite{huang2024can}, the MNIST and CIFAR10 datasets were transformed from traditional image classification datasets into graph representations by constructing an 8-nearest-neighbor graph of SLIC superpixels.     
    \item Global Mask Module: Hyperspectral images are encoded into graph structures by treating each pixel as a node and creating edges based on spatial or spectral similarity, as shown in \cite{yang2024graphmamba}.
    \item 3D Mesh Reconstruction: Hand images can be transformed into graphs by using parametric representations such as MANO to reconstruct 3D hand meshes. Each node represents a key point on the hand, and edges represent connections between these key points, as in \cite{romero2022embodied}.
    
\end{itemize}

\item \textbf{Health Data:}
\begin{itemize}
    \item EEG/MEG Data: Brain activity is represented as a graph, where each channel used for recording (e.g., EEG electrodes) is treated as a node. The edges in this graph are defined by functional connectivity measures such as correlation and phase-locking value (PLV), or imaginary phase-locking value (iPLV) \cite{behrouz2024brain}.
    \item Electronic Health Records (EHRs): Using this type of data, the patients are represented as nodes, and the connections between them are identified based on their shared clinical features or treatment courses.
    \item fMRI: fMRI data is represented as graphs by treating brain regions of interest (ROIs. Edges are defined based on functional connectivity measures across nodes, such as Pearson correlation or coherence. 
\end{itemize}

\item \textbf{Molecular Data:}
\begin{itemize}
    \item Chemical Structures: Molecules are naturally represented as graphs, where atoms serve as nodes and chemical bonds as edges in molecules \cite{han2024innovative}. The graph can be enhanced with advanced representations by leveraging bond types and various atomic properties as features.
    \item Protein Interaction Networks: Similar to molecular structures, proteins can be modeled as graphs, where nodes represent amino acids and edges denote interactions or spatial proximity \cite{han2024innovative}. 
\end{itemize}

\item \textbf{Time-Series Data:}
\begin{itemize}
    \item Dynamic Graphs: Each time step can represent a snapshot where nodes are entities (e.g., sensors in IoT networks), and edges are based on temporal correlations or physical connectivity \cite{li2024dyg, ding2024dygmamba}.
    \item Price Data: Price data can also be transformed into graphs to analyze financial trends and relationships \cite{mehrabian2024mamba}. Nodes in this case represent financial instruments (e.g., stocks, commodities, or cryptocurrencies) or specific time points in a series. Edges capture relationships such as temporal dependencies between consecutive time points.
\end{itemize}


\end{enumerate}

\begin{table}[h]
\centering
\caption{Literature summary of Graph-Mamba tasks, methods and datasets used.}
\label{table:data1}
\resizebox{\textwidth}{!}{%
\begin{tabular}{p{2cm}p{4.5cm}p{7cm}c}
\toprule
\begin{tabular}[c]{@{}l@{}}\textbf{Work} \end{tabular} &  \textbf{Task} &  \begin{tabular}[c]{@{}l@{}}\textbf{Datasets used}\end{tabular}  & \textbf{Code}  \\
\midrule

Huang \textit{et al.} \cite{huang2024can} & Protein classification  &  MNIST, CIFAR10, PATTERN, ZINC, CLUSTER, Ogbg-molhiv, MalNet-Tiny, PascalVOC-SP, Peptides-struct, Peptides-func   & \checkmark \\ \hline
Montagna \textit{et al.} \cite{montagna2024topological} & Simplicial complexes processing & Cora, Citeseer, Pubmed, Minesweeper, Amazon Ratings, Roman Empire, US-county-demos & - \\ \hline
 Lincan \textit{el al.} \cite{li2024stg} &STG prediction & PeMS04, HZMetro, KnowAir  & \checkmark \\ \hline
Li \textit{el al.} \cite{li2024state} & Node classification  & DBLP-3, Brain, Reddit, DBLP-10, ArXiv, Tmall & \checkmark \\ \hline
Choi \textit{el al.} \cite{choi2024spot} & STG forecasting  &  PeMS04 & \checkmark \\ \hline
Tang \textit{el al.} \cite{tang2023modeling} & \makecell[l]{EEG-based seizure detection \\ Diagnosis of sleep disorders \\ ECG classification} & TUSZ, DOD-H, ICBEB ECG & \checkmark \\ \hline
Lawan \textit{el al.} \cite{lawan2024mambaforgcn} & Aspect-based sentiment analysis   & Restaurant14, Laptop14, Twitter & - \\ \hline
Ali \textit{el al.} \cite{mehrabian2024mamba} & Stock price prediction   &  NASDAQ, NYSE, DJIA & \checkmark \\ \hline
Han \textit{el al.} \cite{han2024innovative} & Protein-ligand affinity prediction  &  KIBA, Davis, BindingDB & \checkmark \\ \hline
Xu \textit{el al.} \cite{xu2024identifying} & \makecell[l]{Subphenotypes for sepsis with acute \\ Kidney injury identification } &  MIMIC-IV & - \\ \hline
Pan \textit{el al.} \cite{pan2024hetegraph} & Heterogeneous graph learning  &  ogbn-mag, DBLP, IMDB, ACM &  - \\ \hline
Dong \textit{el al.} \cite{dong2024hamba} & 3D Hand Reconstruction  &  FreiHAND, HO3D, MTC, RHD, InterHand2.6M, H2O3D, DexYCB, COCO-Wholebody, Halpe, MPII NZSL & - \\ \hline
 Zhao \textit{el al.} \cite{zhao2024grassnet} & Node classification   & Cora, CiteSeer, PubMed, Photo, Chameleon, Squirrel, Actor, Texas, Cornell &  - \\ \hline
Yang \textit{el al.} \cite{yang2024graphmamba} & Hyperspectral Image Classification   &  Indian Pines, Salina, UH2013 &  \checkmark \\ \hline
Wang \textit{el al.} \cite{wang2024graph} & \makecell[l]{Classification; Regression; \\ Link ranking}  & Peptides-func, Peptides-struct, PascalVOC-SP, COCO-SP, MalNet-Tiny, PCQM-Contact, CIFAR10, MNIST, CLUSTER, PATTERN &  \checkmark \\ \hline
Zhou \textit{el al.} \cite{zhou2024graph} & Wireless Traffic Prediction &  Telecom Italia dataset &  - \\ \hline
Aghaee \textit{el al.} \cite{aghaee2024graph} &  \makecell[l]{Metabolites’ concentrations\\ prediction\\ Fault Detection and Diagnosis} & Oxidative Stress Metabolic Pathways &  - \\ \hline
 Behrouz \textit{el al.} \cite{behrouz2024graph} & \makecell[l]{Graph classification and regression \\ Node classification \\ Link classification  } &COCO-SP, PascalVOC-SP, Peptides-Func, Peptides-Struct, MNIST, CIFAR10, PATTERN, MalNet-Tiny, Roman-empire, Amazon-ratings, Minesweeper, Tolokers, OGBN-Arxiv  &  \checkmark \\ \hline
Zhang \textit{el al.} \cite{zhang2024gm} & Medical image segmentation & ISIC2017, ISIC2018, Synapse  &  - \\ \hline
Zhou \textit{el al.} \cite{zhou2024efficient} & Pediatric bone age assessment & RSNA, RHPE, DHE & \checkmark \\ \hline
Zhang \textit{el al.} \cite{zhang2024enhanced} &Multi-agent trajectories prediction &  highD, rounD & - \\ \hline
Li \textit{el al.} \cite{li2024dyg} & \makecell[l]{Long-term temporal dependencies \\ capturing on dynamic graphs }   &  Wikipedia, Reddit, MOOC, LastFM, Enron, Social Evo., UCI, Can. Parl., US Legis., UN Trade, UN Vote, Contact & \checkmark \\ \hline
Ding \textit{el al.} \cite{ding2024dygmamba} & \makecell[l]{Dynamic link prediction  \\ Dynamic node classification} &  LastFM, Enron, MOOC, Reddit, Wikipedia, UCI, Social Evo. & - \\ \hline
Ding \textit{el al.} \cite{ding2024combining} & \makecell[l]{Tissue spatial relationships \\ capturing in pathology images}   &  NLST, TCGA  & \checkmark \\ \hline
Behrouz \textit{el al.} \cite{behrouz2024brain} &  Brain activity encoding    & BVFC, BVFC-MEG, ADHD, TUH-EEG, HCP-Age, HCP-Mental, MPI-EEG, TUSZ & \checkmark \\  \hline
\end{tabular}
}
\end{table}

\begin{table}[h]
\centering
\caption{Overview of Different Types of Datasets Used for Graph Mamba Application}
\label{table:datasets1}
\begin{tabular}{lccp{5cm}c}
\toprule
\textbf{Dataset} & \textbf{Modality}  & \textbf{Privacy} & \textbf{Details} & \textbf{Data Access} \\
\midrule
PeMS04 & Graph  & Public & Traffic data, 307 nodes (California) & \href{https://github.com/Davidham3/STSGCN}{URL1}   \\
HZMetro & Graph & Public & Hangzhou metro records & \href{https://github.com/HCPLab-SYSU/PVCGN}{URL2}  \\ 
KnowAir & Graph & Public & Weather records from 184 Chinese cities & \href{https://github.com/shawnwang-tech/PM2.5-GNN}{URL3}  \\
BVFC-MEG & MEG & Public & Magnetoencephalography counterpart of BVFC dataset & \href{https://things-initiative.org/}{URL4} \\
HCP-Mental & Images & Public & 7,440 fMRI samples labeled with 7 mental states & \href{https://www.humanconnectome.org/study/}{URL5} \\
HCP-Age & Images & Public & 7,440 fMRI samples for predicting human subject age & \href{https://www.humanconnectome.org/study/}{URL6} \\
Indian Pines & Image & Public & Hyperspectral dataset, 145×145, 200 bands, 16 classes & \href{https://www.kaggle.com/datasets/abhijeetgo/indian-pines-hyperspectral-dataset}{URL7} \\
Salina & Image & Public & Hyperspectral dataset, 512×217, 204 bands, 16 classes & \href{https://www.kaggle.com/datasets/wangyijialili/salinas/data}{URL8} \\
UH2013 & Image & Public & Hyperspectral dataset, 349×1905, 144 bands, 15 classes & \href{https://github.com/YuxiangZhang-BIT/Data-CSHSI}{URL9} \\
NASDAQ & Price data & Public & Historical daily NASDAQ prices until April 2020 & \href{https://shorturl.at/PvNBv}{URL10} \\
NYSE & Price data & Public & 20-year NYSE trading data (returns, volume, volatility) & \href{https://islp.readthedocs.io/en/latest/datasets/NYSE.html}{URL11} \\
DJIA & Price data & Public & Daily price movements of 30 companies, 2009–2019 & \href{https://shorturl.at/idfTk}{URL12} \\
Restaurant14 & Text & Public & 3,841 English sentences from restaurant reviews & \href{https://shorturl.at/bGkde}{URL13}\\
Laptop14 & Text & Public & 3,845 English sentences from laptop reviews & \href{https://shorturl.at/bGkde}{URL14} \\
Twitter & Text & Public & Data collected via Twitter API & \href{https://x.com/?lang=en}{URL15} \\
\hline
\end{tabular}
\end{table}

\section{Comparative Analysis of Graph Mamba and Recent Alternatives}
\label{Section6new}
In this section, we present a comparative analysis of the performance of Graph Mamba-based approaches against conventional methods, including CNNs, LSTMs, Tranformers, and GNNs.

\subsection{Traffic and Environmental Forecasting}

In the realm of traffic and environmental forecasting, different techniques of ML have been used to predict complex spatiotemporal patterns, including Long Short-Term Memory (LSTM) networks, GNNs, and attention-based architectures. 
Table \ref{tab:TRperformance_comparison} compares the performance of these models, in addition to the Graph Mamba model, on three benchmark datasets: PeMS04 (Flow), HZMetro, and KnowAir. The evaluation metrics include RMSE, MAE, and MAPE.

The performance metrics indicate that attention-based models, particularly STAEformer \cite{liu2023spatio}, outperform LSTM-based and GNN models across all datasets. This suggests that incorporating attention mechanisms enables more effective modeling of complex spatiotemporal dependencies. However, the STG-Mamba model, which leverages a SSM to capture long-range dependencies in spatiotemporal graphs, demonstrates superior predictive accuracy in traffic and environmental forecasting tasks. This enhancement is attributed to Mamba's ability to effectively model intricate spatiotemporal relationships.

\begin{table}[ht!]
\centering
\caption{Performance Comparison of Graph Mamba and State-of-the-Art Models For Traffic Prediction, Urban Flow Forecasting, and Environmental Data }
\label{tab:TRperformance_comparison}
\resizebox{\textwidth}{!}{\begin{tabular}{lccccccccccc}
\hline
\multirow{2}{*}{\textbf{Model}} & \multicolumn{3}{c}{\textbf{PeMS04 (Flow)}} & \multicolumn{3}{c}{\textbf{HZMetro}} & \multicolumn{3}{c}{\textbf{KnowAir}} \\ \cline{2-10} 
 & \textbf{RMSE} & \textbf{MAE} & \textbf{MAPE(\%)}  & \textbf{RMSE} & \textbf{MAE} & \textbf{MAPE(\%)} & \textbf{RMSE} & \textbf{MAE} & \textbf{MAPE(\%)} \\ \hline

\makecell[l]{LSTM-based methods\\ \cite{sutskever2014sequence} \cite{liu2020physical}\cite{tian2024air} } & 40.49 & 26.24 &  19.30  & 57.10 &  35.27 & 9.9 &  30.88 & 20.21 & - \\ 
STGCN \cite{yu2017stgcn} & 35.55 & 22.70 & 14.59  & 34.85 & 21.33 & 13.47  & 11.46 & 8.37 & 11.26\\
STAEformer \cite{liu2023spatio} & 30.18 & 18.22 & 11.98  & 29.94 & 18.85 & 12.03  & 8.69 & 6.93 & 9.89 \\
\rowcolor{lightgray} STG-Mamba \cite{li2024stg}& 29.53 & 18.09 & 12.11  & 29.23 & 18.26 & 11.59 & 8.05 & 6.37 & 9.64 \\ \hline
\end{tabular}}
\end{table}

\subsection{Healthcare and Biosignals}
Grah-Mamba has emerged as a versatile framework with applications across diverse healthcare domains. In the realm of medical image analysis, GM-UNet \cite{zhang2024gm}, a Graph-Mamba-based architecture, has demonstrated promising results in tasks such as skin lesion segmentation and organ segmentation. Extending its reach to pediatric healthcare, GGVMamba \cite{zhou2024efficient} has been successfully employed for pediatric bone age assessment directly from raw X-ray images. In the critical care setting, MGSSM-SAKI \cite{xu2024identifying} leverages Graph-Mamba to identify subphenotypes of SAKI through the analysis of EHR. Furthermore, Graph-Mamba's capabilities extend to computational pathology, as exemplified by GAT-MAMba \cite{ding2024combining}, which accurately predicts progression-free survival among patients with early-stage LUAD based on histopathological images. These applications highlight the versatility and potential of Graph-Mamba in addressing complex challenges within the healthcare field.

While Graph-Mamba has shown promise in various applications, BrainMamba \cite{behrouz2024brain} and GraphS4mer \cite{tang2023modeling} further advance the state-of-the-art in brain classification tasks.
Table \ref{tab:Healthcompare} provides a comparative analysis of these models with baseline methods on a multi-class brain classification task. BrainMamba consistently outperforms all baselines, achieving significant improvements on three datasets (BVFC-MEG, HCP-Mental, HCP-Age). The most substantial improvement of 24.68\% is observed on the Bvfc-MEG dataset, which benefits from BrainMamba's ability to effectively process high-temporal-resolution data. This superior performance can be attributed to BrainMamba's efficiency, scalability, and adaptability to diverse temporality resolutions. Compared to the graph mamba GraphS4mer, BrainMamba is capable of learning from data with a diverse range of temporal resolutions, including high-resolution data like MEG and low-resolution data like fMRI.

\begin{table}[ht!]
\centering
\caption{Performance Comparison of Graph Mamba and State-of-the-Art Models For Multi-class Brain Classification in term of Mean Accuracy (\%) ± std.}
\label{tab:Healthcompare}
\begin{tabular}{lcccc}
\hline
\textbf{Model}     & \textbf{BVFC-MEG}   & \textbf{HCP-Mental} & \textbf{HCP-Age}      \\ \hline
USAD   \cite{audibert2020usad}    & 50.02 $\pm$ 1.13 & 73.49 $\pm$ 1.56 & 39.17 $\pm$ 1.68 \\
GMM    \cite{he2023generalization}    & 53.04 $\pm$ 1.73 & 90.92 $\pm$ 1.83 & 47.75 $\pm$ 1.26 \\
GRAPHMIXER \cite{cong2023we} & 53.12 $\pm$ 1.18 & 91.13 $\pm$ 1.44 & 48.32 $\pm$ 1.11 \\
BRAINNETCNN \cite{kawahara2017brainnetcnn} & 50.12 $\pm$ 1.57 & 83.58 $\pm$ 1.68 & 42.26 $\pm$ 2.03 \\
BRAINGNN \cite{li2021braingnn}   & 51.08 $\pm$ 0.96 & 85.25 $\pm$ 2.17 & 43.08 $\pm$ 1.54 \\
FBNETGEN \cite{kan2022fbnetgen}   & 50.94 $\pm$ 1.39 & 84.47 $\pm$ 1.88 & 42.83 $\pm$ 1.78 \\
ADMIRE  \cite{behrouz2023admire++}    & 54.87 $\pm$ 1.92 & 89.74 $\pm$ 1.93  & 47.82 $\pm$ 1.72 \\
PTGB  \cite{yang2023ptgb}     & 55.11 $\pm$ 1.62 & 92.58 $\pm$ 1.31 & 48.41 $\pm$ 1.47  \\
BNTransformer \cite{kan2022brain} & 55.17 $\pm$ 1.74 & 91.71 $\pm$ 1.48 & 47.94 $\pm$ 1.15 \\
BRAINMIXER \cite{behrouz2023learning} & 62.58 $\pm$ 1.12 & 96.32 $\pm$ 0.29 & 57.83 $\pm$ 1.03 \\
\rowcolor{lightgray} GraphS4mer \cite{tang2023modeling} & 74.39 $\pm$ 0.92 & 82.30 $\pm$ 1.83 & 46.32 $\pm$ 1.09 \\
\rowcolor{lightgray} BrainMamba \cite{behrouz2024brain} & 78.03 $\pm$ 1.69 & 96.57 $\pm$ 1.05 & 59.62 $\pm$ 1.71 \\ \hline
\end{tabular}
\end{table} 

\subsection{Remote Sensing}

Remote sensing image classification is a critical task in geoscience and remote sensing. It provides the accurate interpretation of hyperspectral images to be used in applications such as land use monitoring, agricultural mapping, and urban planning. In this subsection, we evaluate and compare the performance GraphMamba\cite{yang2024graphmamba} with different ML approaches including AB-LSTM \cite{mei2021hyperspectral}, 3D CNN \cite{hamida2018deep}, Transformer GAHT \cite{mei2022hyperspectral}. The used datasets for comparison are: Indian Pines (IP), Salinas (SA), and University of Houston 2013 (UH2013). Table \ref{tab:RScomparison} summarizes the results.

The results highlight the evolution of ML models in remote sensing. AB-LSTM is effective for sequential tasks, but fails to handle the spatial and spectral complexity of hyperspectral data. 3D CNNs address this limitation by incorporating spatial-spectral learning but fall short in capturing global dependencies. Transformer GAHT introduces advanced attention mechanisms to overcome these limitations, but at the cost of computational overhead. GraphMamba combines the strengths of GNNs with spatial-spectral feature learning, delivering state-of-the-art performance with remarkable efficiency. This comparison underscores the importance of using advanced architectures like GraphMamba for hyperspectral image classification, offering a balanced trade-off between accuracy and computational cost.

\begin{table}[ht!]
\caption{Performance Comparison of Graph Mamba and State-of-the-Art Models For Remote Sensing}
\label{tab:RScomparison}\resizebox{\textwidth}{!}{\begin{tabular}{lcccccccccc}

\hline
\textbf{Model/ Data} & \multicolumn{3}{c}{\textbf{Indian Pines}}                                                                                                  & \multicolumn{3}{c}{\textbf{Salina}}                                                                                                  & \multicolumn{3}{c}{\textbf{UH2013}}                                                                                              & \textbf{Complexity} \\ \cline{2-10}
                     & \begin{tabular}[c]{@{}c@{}}\textbf{Overall}\\ \textbf{accuracy}\end{tabular} & \begin{tabular}[c]{@{}c@{}}\textbf{Average}\\  \textbf{accuracy}\end{tabular} & \textbf{Kappa} & \begin{tabular}[c]{@{}c@{}}\textbf{Overall}\\ \textbf{accuracy}\end{tabular} & \begin{tabular}[c]{@{}c@{}}\textbf{Average}\\  \textbf{accuracy}\end{tabular} & \textbf{Kappa} & \begin{tabular}[c]{@{}c@{}}\textbf{Overall}\\ \textbf{accuracy}\end{tabular} & \begin{tabular}[c]{@{}c@{}}\textbf{Average}\\  \textbf{accuracy}\end{tabular} & \textbf{Kappa} &\textbf{ (G)}                 \\ \hline
AB-LSTM \cite{mei2021hyperspectral}              & 56.68                                                      & 60.41                                                       & 51.02 & 74.83                                                      & 83.89                                                       & 72.21 & 79.61                                                      & 82.38                                                       & 77.97 & -                   \\ \hline
3D CNN \cite{hamida2018deep}               & 65.83                                                      & 79.12                                                       & 61.38 & 87.59                                                      & 92.96                                                       & 86.21 & 71.84                                                      & 73.84                                                       & 69.54 & 0.69                \\ \hline
\makecell[l]{Transformer\\ GAHT \cite{mei2022hyperspectral}}     & 85.96                                                      & 92.96                                                       & 84.05 & 94.21                                                      & 97.37                                                       & 93.56 & 90.84                                                      & 92.13                                                       & 90.10 & 4.27                \\ \hline
\rowcolor{lightgray} \makecell[l]{GraphMamba \\\cite{yang2024graphmamba}    }       & 96.43                                                      & 97.83                                                       & 95.91 & 97.33                                                      & 98.66                                                       & 97.02 & 95.62                                                      & 96.23                                                       & 95.27 & 069                 \\ \hline
\end{tabular}}
\end{table}

\subsection{Financial Markets and Economic Forecast}
The task of financial market forecasting involves predicting the future behavior of key indices. These predictions are crucial for different tasks such as decision-making, risk management, and economic planning. State-of-the-art ML models have been used for this purpose, including LSTM, Transformer-based architectures, GNNs, and hybrid models. In this section, we compare the performance of Graph Mamba (SAMBA) with other baseline models, focusing on three key evaluation metrics: RMSE (Root Mean Square Error), IC (Information Coefficient), and RIC (Ranked Information Coefficient). The results are summarized in Table~\ref{tab:FMcomparison}.

The outstanding performance of SAMBA can be attributed to its graph-based design, which allows for the modeling of complex relationships between financial indices. Unlike LSTM and Transformer models, SAMBA's ability to leverage graph structure provides a more comprehensive representation of multivariate time series data. Moreover, its efficient architecture ensures that the additional computational overhead is minimal compared to the Transformer model. 

\begin{table}[ht!]
\centering

\caption{Performance Comparison of Graph Mamba and State-of-the-Art Models  For Financial Market Forecasting} \label{tab:FMcomparison}
\begin{tabular}{lcccc}
\hline
\multicolumn{1}{r}{\textbf{Metric}}                                                    & \multicolumn{3}{c}{\textbf{RMSE}}               & \textbf{Training Time} \\ \cline{2-4}
\textbf{Model/ Data}                                                                   & \textbf{NASDAQ} & \textbf{NYSE} & \textbf{DJIA} & \textbf{second/epoch}  \\ \hline
LSTM \cite{moghar2020stock}                                           & 0.0187          & 0.0144        & 0.0121        & 0.081                  \\ \hline
Transformer \cite{wang2022stock}                                      & 0.0147          & 0.0141        & 0.0166        & 1.91                   \\ \hline
FourierGNN \cite{yi2024fouriergnn}                                    & 0.0152          & 0.0146        & 0.019         & 1.10                   \\ \hline
\rowcolor{lightgray} SAMBA \cite{mehrabian2024mamba} & 0.0128          & 0.0125        & 0.0108        & 0.891                  \\ \hline
\end{tabular}
\end{table}

\subsection{Sentiment Analysis}
In this section, we present a comparative analysis of the MambaForGCN model against state-of-the-art architectures, including CNN, GCN, and transformer-based architectures. Three benchmark datasets are used for evaluation, including: Restaurant14, Laptop14, and Twitter, with Accuracy and F1-score as the performance metrics. The results, detailed in Table \ref{tab:SAperformance}, demonstrate that MambaForGCN consistently surpasses existing models across all datasets.

The MambaForGCN model outperforms other architectures because of its innovative integration of syntactic and semantic information. It is able to effectively capture both short and long range dependencies between aspect and opinion words. Traditional models like CNNs and GCNs fail with long-range dependencies and complex syntactic structures, while Transformer-based models such as the dependency graph enhanced dual-transformer network (DGEGT) face challenges with quadratic complexity in attention mechanisms that limit their efficiency in handling extensive dependencies. MambaForGCN addresses these limitations by integrating graph learning with a Mamba module. This combination allows the model to capture both syntactic structures and semantic information and enhance its ability to understand complex relationships between aspects and opinions in text. 

\begin{table}[ht!]
\centering
\caption{Performance Comparison of Graph Mamba and State-of-the-Art Models  For Sentimental Analysis}
\label{tab:SAperformance}
\resizebox{\textwidth}{!}{\begin{tabular}{lcccccc}
\hline
\textbf{Model/ Data}        & \multicolumn{2}{c}{\textbf{Restaurant14}} & \multicolumn{2}{c}{\textbf{Laptop14}} & \multicolumn{2}{c}{\textbf{Twitter14}} \\ \cline{2-7} 
                      & \textbf{Accuracy}    & \textbf{F1-score}           & \textbf{Accuracy}   & \textbf{F1-score}         & \textbf{Accuracy}   & \textbf{F1-score}       \\ \hline

CNN Over BERT-GCN \cite{phan2022aspect}  & 85.25& 78.76 &78.59&74.76&74.03 &72.39 \\\hline
Dual-gated GCN \cite{liu2023enhancing} & 83.66 &76.73& 75.70& 72.57& 74.87& 72.27 \\\hline
\makecell[l]{DGEDT \cite{tang2020dependency}}  & 83.90 &75.10 &76.80 &72.30 &74.80 &73.40 \\ \hline
\rowcolor{lightgray} MambaForGCN \cite{liu2024kan}& 86.68& 80.86& 81.80& 78.59& 77.67& 76.88
            \\ \hline
\end{tabular}}
\end{table}

\section{Challenges in Applying Graph Mamba}
\label{sec6}

Graph Mamba models represent a significant evolution in graph learning, taking advantage of state-space models to manage both the complexity and dynamism of graph-structured data. However, like many advanced models, Graph Mamba faces several application challenges, particularly with regard to scalability, interpretability, and training convergence. These challenges are magnified in the face of increasingly large and heterogeneous graph datasets. In the following, we discuss these core challenges and examine emerging trends and potential solutions.

\subsection{Scalability}
Scalability remains one of the foremost challenges in applying Graph Mamba to real-world data, where graphs often contain millions or billions of nodes and edges. This complexity increases both computational and memory demands, posing difficulties in maintaining efficient performance and accuracy.

\subsubsection{Graph Size and Complexity}
Processing large graphs requires the handling of extensive node and edge relationships, resulting in an exponential growth of computations as the size of the graph increases. Traditional methods have issues with memory overflow and slower processing times. To address these issues, techniques such as node sampling and edge pruning have been employed to selectively process relevant portions of the graph, therefore reducing computational load. However, these methods are susceptible to losing important connections that affect the accuracy of the model. 
Recent advancements in adaptive sampling methods, which dynamically adjust based on data, offer a promising solution by balancing computational efficiency with the retention of essential structural information.

\subsubsection{Computational Efficiency} Although state-space models within Graph Mamba reduce complexity, further efficiency is needed to handle massive graph datasets. High memory consumption due to multi-layered architectures exacerbates this issue. Advances in specialized hardware, such as Graph Processing Units (GPUs) and Field-Programmable Gate Arrays (FPGAs), are advancing computational efficiency by offering parallel processing capabilities suited to large-scale graph learning. Additionally, the exploration of graph-based data compression techniques, such as dimensionality reduction and sparse matrix representations \cite{liu2020discriminative}, is emerging as a promising approach to alleviate the limitations of hardware and memory resources. These innovations collectively pave the way for more scalable and efficient graph processing frameworks.

\subsubsection{Edge Sparsity and Dense Clusters} Large graphs can be sparse, with varying connectivity between nodes. Dense clusters and highly connected subgraphs add to processing time and storage complexity.
Using sparse matrices and dynamic pruning can reduce memory requirements by focusing on meaningful edges \cite{hu2023dynamic,gurevin2024prunegnn}. Furthermore, graph partitioning methods facilitate localized processing by dividing the graph into smaller and manageable components. These components can be processed independently and then recombined. This can effectively address the challenges posed by both sparse and dense regions while optimizing computational efficiency.

\subsection{Interpretability}
The interpretability of Graph Mamba models is essential for practical applications, especially in critical fields like healthcare, finance, and social sciences. However, as Graph Mamba integrates DL with state-space models, its complexity makes it challenging to interpret the decision-making process.

\subsubsection{Layer Complexity and Black-Box Nature}
Graph Mamba models comprise multiple layers of state-space transformations, making it difficult to trace how node and edge attributes influence final predictions. This lack of transparency is particularly concerning in fields that require accountability, like healthcare or finance.
Recent trends focus on attention mechanisms and visualization techniques that highlight important nodes and edges involved in predictions \cite{feng2022kergnns}. For instance, node-level attention weights can reveal which connections and features contribute most significantly to the outcome, improving transparency.

\subsubsection{Complexity of Message Passing Mechanisms}
The complex message-passing architectures within Graph Mamba models, designed to handle complex relationships, can make it challenging to interpret inter-node interactions.
To address these challenges, eXplainable AI (XAI) techniques for GNNs are being adapted to state-space models, providing interpretable outputs. Tools like GraphLIME and GNNExplainer are being tailored for Graph Mamba, enabling insights into message-passing pathways and making the model’s reasoning more accessible.

\subsubsection{Model-Specific Interpretability for Domain-Specific Applications}
Certain applications demand tailored interpretability features to meet the unique requirements of their respective domains. For instance, in healthcare, models may need to highlight biological pathways or critical nodes influencing patient outcomes, while in social media analytics, the focus might be on tracing network interactions and community dynamics. Domain-adaptive explainability methods are emerging as a valuable approach, where models provide outputs relevant to specific fields \cite{chaddad2023explainable}. For example, healthcare-focused Graph Mamba models could be designed to emphasize pathways or critical nodes associated with patient outcomes, making the model's decisions more understandable and relevant to medical professionals.

\subsection{Training and Convergence}
Efficient training and convergence of Graph Mamba are critical, especially for highly heterogeneous graphs that span across different domains and have nodes with diverse attributes. Ensuring stable and efficient training on such data poses significant challenges.
\subsubsection{Slow Convergence in Heterogeneous Graphs}
Heterogeneous graphs contain nodes and edges of varying types, making it challenging for Graph Mamba to converge quickly and accurately. Each node type requires different feature transformations which increases the complexity of training.
To address this challenge, heterogeneous normalization layers and adaptive learning rates \cite{spinelli2020adaptive} can be used to speed up the convergence by treating different node and edge types individually. This approach improves model adaptability and allows each part of the graph to converge at an optimal rate.

\subsubsection{Computational Cost of Training} The extensive training cycles required for Graph Mamba models, particularly on large and complex datasets, present significant computational demands. Training large-scale graphs often involves billions of computations per epoch, which makes the process time-intensive.
In this case, incremental and online learning techniques, such as mini-batch training and graph sampling \cite{yang2023batchsampler}, allow the model to update in response to new data without retraining entirely from scratch. 

\subsubsection{Training on Temporal and Evolving Graphs} 
Dynamic and temporal graphs that evolve over time present special training challenges, as node attributes and connections change, requiring the model to adapt constantly.
Promising solutions include continual learning and recurrent state-space models, which enable efficient updates by incorporating new data incrementally without the need for retraining the entire model. Additionally, transfer learning techniques facilitate the reuse of knowledge from previously trained models, allowing Graph Mamba to dynamically integrate new information while preserving insights from past data. These methods allow Graph Mamba to incorporate new information dynamically, improving model flexibility and adaptability to evolving data.
 
\section{Future Research Directions}
\label{sec7}

Several future research perspectives on Graph Mamba models could be explored to address emerging challenges and align with recent technological trends. In the following, we discuss several promising directions.

\subsection{Improving the efficiency of the model}
Graph Mamba models, while powerful, face challenges in scaling to handle extensive and complex graph data. As the size of datasets increases, so does the demand for computational resources, which can limit the model’s real-world applicability, especially in high-demand settings like social media analysis. Future research could focus on optimizing these models to be more computationally efficient by:
\begin{itemize}
    \item \textbf{Algorithmic Optimizations:} By reducing the layers of computation and focusing only on the most relevant nodes and edges, models can be streamlined for faster processing without sacrificing accuracy.
    \item \textbf{Hardware Integration:} Leveraging specialized hardware like GPUs could boost efficiency and scalability, especially for tasks involving high-volume, complex graph data. Further exploration in designing custom chips optimized for state-space models could push efficiency gains even further.
    \item \textbf{Memory Management Techniques:} Developing advanced memory management strategies could allow Graph Mamba models to process larger graphs with limited hardware resources, making them more accessible and practical across different fields.
\end{itemize}

These improvements will be crucial in making Graph Mamba more accessible and viable for a wider range of applications. They will enable real-time analysis in domains that demand processing at scale.

\subsection{Progress in SSL}
Traditional graph-based models often require extensive labeled data for training, which can be both costly and time-consuming to produce. SSL presents a promising alternative by allowing models to learn from unlabeled data, thus making it feasible to train on larger and more diverse datasets \cite{liu2022graph,chong2020graph}. Future research could investigate how SSL techniques can be tailored to enhance Graph Mamba models:
\begin{itemize}
    \item \textbf{Graph-Specific SSL Tasks:} Designing pretext tasks specific to graph data, such as predicting node or edge attributes, could allow the model to uncover relationships and patterns without explicit labels. This will enhance the model's capacity to generalize across different graph structures.
    \item \textbf{Improving Feature Extraction: }SSL could enable Graph Mamba to automatically identify relevant features across various domains, such as node interactions in social networks or protein structures in bioinformatics.
\item \textbf{Reduced Label Dependency:} By employing unlabeled data, SSL in Graph Mamba could improve performance in domains where annotated data is sparse or expensive, such as biological networks and evolving social structures.
\end{itemize}
Incorporating SSL into Graph Mamba will allow for faster scaling of research and applications without needing extensive labeling efforts, thereby extending the bounds of graph-based learning.

\subsection{Dynamic Graph Adaptability}
Many real-world applications involve graphs that evolve over time, such as financial transactions and molecular interactions \cite{feng2024dynamicgnns}. These dynamic graphs change in structure and attribute values, presenting challenges for traditional graph models that assume static structures. Research in dynamic graph adaptability could explore:

\begin{itemize}
    \item \textbf{Real-Time State-Space Updates:} Developing state-space models that can update in real-time as the graph evolves would allow Graph Mamba to reflect current structures and interactions and capture temporal dependencies more effectively.
\item \textbf{Temporal and Spatial Awareness:} Enhancing Graph Mamba’s ability to handle both spatial (network structure) and temporal (changes over time) dependencies could lead to models that are more robust to time-dependent data changes. Thi will improve the predictive accuracy in different domains like traffic prediction and disease progression.

\item \textbf{Incremental Learning Mechanisms:} Implementing incremental learning capabilities would allow Graph Mamba to adapt to changes in the data without the need to retrain the entire model. This would be especially valuable in applications where rapid responses are required, such as monitoring financial markets or social network behavior.
\end{itemize}
Future Graph Mamba models could enable more responsive and accurate analysis in fields requiring real-time data insights by focusing on dynamic graph adaptability.

\subsection{Cross-Domain and Multi-Modal Integration}
While Graph Mamba excels at handling graph data, many real-world scenarios benefit from combining multiple data types, such as images, text, and numerical data, alongside graph structures \cite{zou2025deepfusion,chen2024knowledgegraphs}. Cross-domain and multi-modal integration involve creating models that can effectively learn from multiple sources of information \cite{shi2024graph}. Future research could focus on:
\begin{itemize}
    \item \textbf{Hybrid Models for Multi-Modal Learning:} Building models that incorporate graphs with other modalities, such as images in medical diagnosis or text in social network analysis, could provide a more holistic view and improve model robustness.
\item \textbf{Unified Feature Spaces:} Creating a unified representation of data from diverse domains would enable Graph Mamba to draw insights from multiple types of data simultaneously and improve accuracy in cross-domain tasks such as event prediction or trend analysis.
\item \textbf{Application to Interdisciplinary Fields:} Multi-modal Graph Mamba models could transform fields like healthcare, where integrating electronic health records, medical imaging, and genetic data into a single predictive model could greatly enhance diagnostic capabilities.
\end{itemize}
Cross-domain integration expands the potential applications of Graph Mamba by allowing it to handle more complex and layered data, leading to richer insights than those derived from a single data type.

\subsection{Heterogeneous Graph Handling}
In many applications, graphs are not uniform and contain various types of nodes and edges representing different entities and relationships. These heterogeneous graphs are especially prevalent in social networks, recommendation systems, and biological networks, where different entities and interactions coexist \cite{bing2023heterogeneous}. Research in this area could focus on:
\begin{itemize}
    \item \textbf{Adaptable Node and Edge Representations:} Developing flexible representation techniques that account for multiple node types and interactions could allow Graph Mamba to better capture the complexity of heterogeneous graphs, enhancing model expressiveness and prediction power.
\item \textbf{Customized Message Passing Mechanisms:} By adjusting message passing for different node types and edge relationships, the model could better capture the unique interactions within heterogeneous data structures and enable more accurate inferences.
\item \textbf{Application to Knowledge Graphs and Complex Systems:} Future Graph Mamba models designed for heterogeneous graphs could have significant applications in knowledge graphs, where entities like people, organizations, and concepts interconnect. This would be beneficial in recommendation systems, where different user-item relationships require tailored processing.
\end{itemize}
Supporting heterogeneous graph structures would broaden Graph Mamba’s applicability across fields where data diversity and complexity are inherent, leading to models that offer richer, more reliable insights in applications like personalized recommendations or knowledge discovery.

\section{Conclusion}
\label{sec8}

This survey highlights the impressive progress that Graph Mamba models have made by integrating SSMs into graph-based learning. These models address significant challenges faced by traditional GNNs, particularly by improving efficiency, scalability, and adaptability to complex and dynamic graph structures. Graph Mamba has demonstrated effectiveness across various domains, including recommendation systems and healthcare applications. Key strengths of Graph Mamba models include their ability to capture long-range dependencies, process heterogeneous graph structures, and handle evolving data in real-time.

The potential applications of Graph Mamba models are vast, extending to any field that deals with complex, structured, and changing data. Practitioners can leverage these models to extract deeper insights from graph-based data, potentially enhancing decision-making in areas such as social behavior analysis, personalized recommendations, and medical diagnosis. For researchers, Graph Mamba opens new avenues for further exploration, particularly in the areas of self-supervised learning, dynamic adaptability, and cross-domain data integration. Developing solutions that enhance model interpretability, improve training efficiency and ensure scalability across large datasets will be essential for maximizing Graph Mamba’s utility.


 \bibliographystyle{elsarticle-num} 
 \bibliography{cas-refs}

\end{document}